\documentclass[10pt,twocolumn,letterpaper]{article}

\usepackage{wacv}
\usepackage{times}
\usepackage{epsfig}
\usepackage{graphicx}
\usepackage{amsmath}
\usepackage{amssymb}

\usepackage{color}
\usepackage{float}
\usepackage{multirow}
\usepackage{graphicx}
\usepackage{subfigure}
\usepackage{subfiles}
\usepackage{algorithm}
\usepackage{algpseudocode}

\usepackage{amsmath,amsfonts,bm}

\def\eqref#1{equation~\ref{#1}}
\def\1{\bm{1}}

\DeclareMathAlphabet{\mathsfit}{\encodingdefault}{\sfdefault}{m}{sl}
\SetMathAlphabet{\mathsfit}{bold}{\encodingdefault}{\sfdefault}{bx}{n}

\DeclareMathOperator*{\argmax}{arg\,max}

\usepackage[pagebackref=true,breaklinks=true,letterpaper=true,colorlinks,bookmarks=false]{hyperref}

\wacvfinalcopy
\def\wacvPaperID{69}

\begin{document}

\title{Generalized Clustering and Multi-Manifold Learning with \\ Geometric Structure Preservation}

\author{Lirong Wu {$^{1,2}$}, Zicheng Liu {$^{2}$}, Jun Xia {$^{2}$}, Zelin Zang {$^{2}$}, Siyuan Li {$^{2}$}, Stan Z. Li {$^{2,^\dagger}$} \\
	$^1$ Zhejiang University \quad $^2$  Westlake University \\
	{\tt\small \{wulirong, liuzicheng, xiajun, zangzelin, lisiyuan, stan.zq.li\}@westlake.edu.cn}
}

\maketitle

\begin{abstract}
Though manifold-based clustering has become a popular research topic, we observe that one important factor has been omitted by these works, namely that the defined clustering loss may corrupt the local and global structure of the latent space. In this paper, we propose a novel Generalized Clustering and Multi-manifold Learning (GCML) framework with geometric structure preservation for generalized data, i.e., not limited to 2-D image data and has a wide range of applications in speech, text, and biology domains. In the proposed framework, manifold clustering is done in the latent space guided by a clustering loss. To overcome the problem that the clustering-oriented loss may deteriorate the geometric structure of the latent space, an isometric loss is proposed for preserving intra-manifold structure locally and a ranking loss for inter-manifold structure globally. Extensive experimental results have shown that GCML exhibits superior performance to counterparts  in terms of qualitative visualizations and quantitative metrics, which demonstrates the effectiveness of preserving geometric structure. Code has been made available at: \url{https://github.com/LirongWu/GCML}.
\end{abstract}

\vspace{-1.5em}
\section{Introduction}
Clustering, a fundamental tool for data analysis and visualization, has been an essential research topic in data science. This paper focuses on \emph{generalized clustering}, which takes vector data as input and is applicable to data with various dimensions, not limited to 2-D image data. Conventional clustering algorithms such as $K$-Means \cite{macqueen1965some}, Gaussian Mixture Models \cite{bishop2006pattern}, and Spectral Clustering \cite{shi2000normalized} perform clustering based on distance or similarity measures. However, handcrafted distance or similarity measures are rarely reliable for large-scale high-dimensional data, making it increasingly challenging to achieve effective clustering. An intuitive solution is to transform the data from the high-dimensional input space to the low-dimensional latent space and then to cluster the data in the latent space. This can be achieved by applying manifold-based dimensionality reduction techniques, such as t-SNE \cite{maaten2008visualizing}, and UMAP \cite{mcinnes2018umap}. However, since these methods are not specifically designed for clustering tasks, some of their properties may be contrary to our expectations, e.g., two data points from different manifolds that are close in the input space will be closer in the latent space learned by UMAP. Therefore, the first question here is \emph{how to perform multi-manifold learning for dimensionality reduction that favors clustering?}

The two main points for the multi-manifold learning are \textit{Point (1)} preserving the local geometric structure within each manifold and \textit{Point (2)} ensuring the discriminability between different manifolds. Most previous works seem to start with the assumption that data labels are known, and then design the algorithm in a \emph{supervised manner}. However, it is challenging to decouple complex crossover relations and ensure discriminability between different manifolds, especially in \emph{unsupervised} settings. One natural strategy is to achieve \textit{Point (2)} through clustering to get pseudo-labels and then performing single-manifold learning for each manifold. However, the clustering-oriented loss may deteriorate the geometric structure of the latent space\footnote{This claim was first made by IDEC \cite{guo2017improved}, but they did not provide any experiment or analysis to support it. In this paper, however, we show that the geometric structure of the latent space is indeed corrupted by extensive visualizations (Fig.~\ref{fig:3} and Fig.~\ref{fig:4}) and statistical analysis (Fig.~\ref{fig:6}).}, and hence clustering is somewhat contrary to \textit{Point (1)} (this will be detailed in Sec.~\ref{sec:3.3}). Therefore, it is important to alleviate this contradiction so that clustering helps both \textit{Point (1)} and \textit{Point (2)}. Thus, the second question here is \emph{how to cluster data that favors multi-manifold learning?}

In this paper, we propose to jointly perform generalized clustering and multi-manifold learning with geometric structure preservation. Inspired by \cite{xie2016unsupervised}, the clustering centers are defined as a set of \emph{\textbf{learnable}} parameters, and we use a clustering loss to simultaneously guide the separation of data points from different manifolds and the learning of the clustering centers. To prevent clustering loss from deteriorating the latent space, an isometric loss and a ranking loss are proposed to preserve the intra-manifold local structure and inter-manifold global structure. Finally, we achieve the following three goals related to clustering, geometric structure preservation, and multi-manifold learning: (1) Clustering helps to ensure inter-manifold discriminability \emph{(Point 2)}; (2) Local structure preservation \emph{(Point 1)} can be compatible with clustering; (3) Geometric structure preservation helps to cluster. The contributions are summarized as:

\begin{itemize}
\item Proposing to combine generalized deep clustering and multi-manifold learning into a unified framework with local and global structure preservation.
\item Setting the clustering centers as a set of \emph{learnable} parameters and achieve global structure preservation in a faster, more efficient, and easier to optimize manner by applying ranking loss to the clustering centers.
\item Pointing out contradictions between two optimization goals of clustering and local structure preservation and proposing an elegant training strategy to alleviate it.
\end{itemize}

\begin{figure*}[!htbp]
	\begin{center}
		\includegraphics[width=0.7\linewidth]{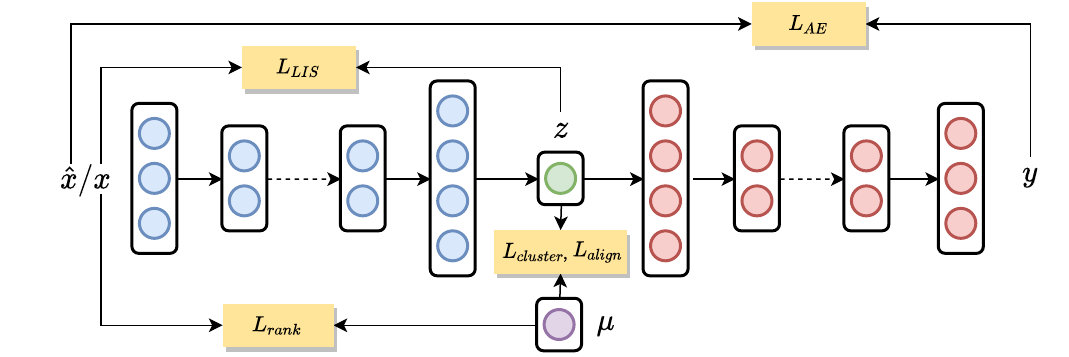}
	\end{center}
	\vspace{-1em}
	\caption{The GCML framework. The encoder, decoder, latent space, and cluster centers are marked as blue, red, green, and purple.}
	\vspace{-1em}
	\label{fig:1}
\end{figure*}

\vspace{-1.5em}
\section{Related Work}
\vspace{-0.5em}
\textbf{Clustering Analysis.} As a fundamental tool in machine learning, clustering has been widely applied in various domains. One branch of classical clustering is $K$-Means and Gaussian Mixture Models (GMM), which are fast, easy to understand, and can be applied to a large number of problems. However, limited by Euclidean measures, their performance on high-dimensional data is often unsatisfactory. Spectral clustering and its variants (such as SC-Ncut \cite{bishop2006pattern}) extend clustering to high-dimensional data by allowing more flexible similarity measures. However, limited by the computational efficiency of the full Laplace matrix, spectral clustering is challenging to extend to large-scale datasets.

\textbf{Deep Clustering.} The success of deep learning has contributed to the growth of \textit{deep clustering} \cite{gao2020clustering,li2020consistent}. One branch of deep learning performs clustering after learning low-dimensional embeddings. For example, \cite{tian2014learning} uses autoencoder to learn low-dimensional features and then runs $K$-Means to get clustering results (AE+$K$-Means). Instead, N2D \cite{mcconville2019n2d} applies UMAP to find the best clusterable manifold of the learned embeddings, and then runs $K$-Means to discover higher-quality clusters. The other category of algorithms tries to optimize clustering and dimensionality reduction jointly. The closest work to us is DEC \cite{xie2016unsupervised}, which learns a mapping from the input space to a low-dimensional latent space through iteratively optimizing clustering-oriented objective. As a modified version of DEC, while IDEC claims to preserve the local structure of data, their contribution is nothing more than adding a reconstruction loss. Besides, JULE \cite{yang2016joint} unifies representation learning with clustering, which can be considered as a neural extension of hierarchical clustering. Instead, SpectralNet \cite{shaham2018spectralnet} directly embeds the input into the Laplacian eigenspace in which clustering is performed. DSC \cite{yang2019deep} devises a dual autoencoder to embed data into latent space, and then deep spectral clustering is applied to obtain label assignments. Moreover, DDC \cite{ren2020deep} proposes to use a density-based clustering algorithm to initialize cluster centers and then perform image cluster discovery.

To avoid any possible misunderstanding, we would like to highlight that \emph{generalized deep clustering and visual self-supervised learning (SSL)} are two different research fields. SSL typically uses more powerful CNN architecture (applicable only to 2-D image data) and applies sophisticated techniques such as contrastive learning \cite{he2020momentum}, data augmentation, and clustering \cite{zhan2020online,van2020scan} for better performance on large-scale datasets, such as ImageNet. For example, ASPC-DA \cite{guo2019adaptive} combines data augmentation with self-paced learning to encourage the learned embeddings to be cluster-oriented. Besides, ClusterGAN \cite{mukherjee2019clustergan} trains a generative adversarial network jointly with a clustering-specific loss to achieve clustering in the latent space. The generalized deep clustering, on the other hand, uses a generalized MLP architecture (applicable to all kinds of data with various dimensions, not limited to 2-D image data) and has a very wide range of applications in images, text, and biology domains.

\textbf{Manifold Learning.} The manifold assumption hypothesizes that a low-dimensional manifold is embedded in a high-dimensional space, and the manifold learning aims to achieve dimensionality reduction via a nonlinear mapping that preserves the geometric structure. Isomap, as a classical algorithm of \emph{single-manifold} learning, aims to seek a global optimal subspace that best preserves the geodesic distance between data points \cite{tenenbaum2000global}. In contrast, some algorithms, such as the Locally Linear Embedding (LLE) \cite{roweis2000nonlinear}, are more concerned with the preservation of local neighborhood information. Furthermore, \emph{multi-manifold} learning has been proposed to obtain intrinsic properties of different manifolds. \cite{yang2016multi} proposes a supervised discriminant isomap where data points are partitioned into different manifolds according to label information. Similarly, \cite{zhang2018semi} proposes a semi-supervised learning framework that applies the labeled and unlabeled training samples to perform the joint learning of local-preserving features. In most previous work, it is assumed that the label is known or partially known, which significantly simplifies the problem. However, it is challenging to decouple multiple overlapping manifolds in unsupervised settings, and that is what this paper aims to explore.

\vspace{-1em}
\section{Proposed Method}
\vspace{-0.5em}
Consider a dataset $X$ with $N$ samples, and each sample $x_i \in \mathbb{R}^{d}$ is sampled from $C$ different manifolds $\left\{M_{c}\right\}_{c=1}^{C}$. Assume that each category in the dataset lies in a compact low-dimensional manifold, and the number of manifolds $C$ is prior knowledge. Define two nonlinear mapping $z_i=f(x_i,\theta_f)$ and $y_i=g(z_i,\theta_g)$, where $z_i \in \mathbb{R}^{m}$ is the embedding of $x_i$ in the latent space, and $y_i$ is the reconstruction of $x_i$. The $j$-th cluster center is denoted as $\mu_j \in \mathbb{R}^{m}$, where $\left\{\mu_{j}\right\}_{j=1}^{C}$ is defined as a set of \emph{\textbf{learnable}} parameters. We aim to find optimal parameters $\theta_f$ and $\left\{\mu_{j}\right\}_{j=1}^{C}$ so that the embeddings $\left\{z_{i}\right\}_{i=1}^{N}$ can achieve clustering with local and global structure preservation. To this end, a denoising autoencoder shown in Fig.~\ref{fig:1} is first pre-trained in an unsupervised manner to learn an initial latent space. Denoising autoencoder aims to optimize the self-reconstruction loss $L_{AE}=MSE(\hat{x},y)$, where the $\hat{x}$ is a copy of $x$ with Gaussian noise added, that is, $\hat{x}=x+N(0,\sigma^2)$. Then the autoencoder is fine-tuned by optimizing the following clustering-oriented loss $\{L_{cluster}(z, \mu)\}$ and structure-oriented losses $\{L_{rank}(x, \mu),L_{\textit{LIS}}(x, z),L_{align}(z, \mu)\}$. Since the clustering should be performed on features of \emph{clean} data, instead of noised data $\hat{x}$ that is used in denoising autoencoder, the clean data $x$ is used for fine-tuning.

\subsection{Clustering-oriented Loss}
First, the cluster centers $\left\{\mu_{j}\right\}_{j=1}^{C}$ in the latent space $Z$ are initialized (the initialization method will be introduced in Sec.~\ref{sec:4.1}). Then the similarity between the embedded point $z_i$ and cluster centers $\left\{\mu_{j}\right\}_{j=1}^{C}$ is measured by Student’s $t$-distribution, as follows
\begin{equation}
	q_{ij}=\frac{\left(1+\left\|z_{i}-\mu_{j}\right\|^{2}\right)^{-1}}{\sum_{j^{\prime}}\left(1+\left\|z_{i}-\mu_{j^{\prime}}\right\|^{2}\right)^{-1}}
\end{equation}
\indent The auxiliary target distribution is further designed to help manipulate the latent space, defined as:
\begin{equation}
	p_{i j}=\frac{q_{i j}^{2} / f_{j}}{\sum_{j^{\prime}} q_{i j^{\prime}}^{2} / f_{j^{\prime}}},\quad where\quad f_{j}=\sum_{i} q_{i j}
\end{equation}
where $f_j$ is the normalized cluster frequency, used to balance the size of different clusters. Then the encoder is optimized by the following objective:
\begin{equation}
	L_{cluster}=\mathrm{KL}(P \| Q)=\sum_{i} \sum_{j} p_{i j} \log \frac{p_{i j}}{q_{i j}}
\end{equation}
\indent The gradient of $L_{cluster}$ with respect to each learnable cluster center $\mu_{j}$ can be computed as:
\vspace{-0.5em}
\begin{small}
\begin{equation}
	\begin{aligned}\frac{\partial L_{cluster}}{\partial \mu_{j}}=-\sum_{i}\left(1+\left\|z_{i}-\mu_{j}\right\|^{2}\right)^{-1} \cdot \left(p_{i j}-q_{i j}\right)\left(z_{i}-\mu_{j}\right)\end{aligned}
\end{equation}
\end{small}

\vspace{-1.5em}
\noindent where $L_{cluster}$ facilitates the aggregation of data points within the same manifold, while data points from different manifolds are kept away from each other. However, we found that the clustering-oriented losses may deteriorate the geometric structure of the latent space. To prevent this deterioration, we introduces three other structure-oriented losses to preserve the local and global manifold structures.

\subsection{Structure-oriented Loss}
\textbf{Intra-manifold Isometry Loss.}
The intra-manifold local structure is preserved by optimizing the objective as:
\vspace{-0.5em}
\begin{small}
\begin{equation}
L_{\textit{LIS}}=\sum_{i=1}^{N} \sum_{j \in \mathcal{N}^Z_{i}}\left|d_X\left(x_i, x_j\right)- d_Z\left(z_i, z_j\right)\right|\cdot\pi(l(x_i)=l(x_j))
\end{equation}
\end{small}

\vspace{-1.5em}
\noindent where $\mathcal{N}^Z_{i}$ represents the neighborhood of data point $z_i$ in the latent space $Z$, and the $k$NN is applied to determine the neighborhood. $\pi(\cdot)\in\{0,1\}$ is an indicator function, and $l(x_i)$ is a \emph{manifold determination function} that returns the manifold $s_i$ where sample $x_i$ is located, that is, $s_i=l(x_i)=\argmax_{j}p_{ij}$. Then we can derive $C$ manifolds $\left\{M_{c}\right\}_{c=1}^{C}$ by $M_{c}=\{x_i;\ s_i=c,i=1,2,...,N\}$. The loss $L_{\textit{LIS}}$ constrains the isometry within each manifold.

\textbf{Inter-manifold Ranking Loss.}
The inter-manifold global structure is preserved by optimizing the objective as:
\vspace{-0.5em}
\begin{equation}
	L_{rank}=\sum_{i=1}^{C} \sum_{j=1}^{C}\left|d_Z\left(\mu_i, \mu_j\right)-\kappa \cdot d_X\left(v^X_i, v^X_j\right)\right|
\end{equation}
where $\{v^X_j\}_{j=1}^{C}$ is defined as the ground-truth centers of different manifolds in the input space $X$ with $v^X_j=\frac{1}{|M_j|}\sum_{i\in M_j}x_i$ $(j=1,2,...,C)$. The parameter $\kappa$ determines the extent to which different manifolds move away from each other. The larger $\kappa$ is, the further away the different manifolds are from each other. The derivation for the gradient of $L_{rank}$ with respect to each learnable cluster center $\mu_{j}$ is placed in \textbf{Appendix A.2}. Note that $L_{rank}$ is optimized in an \emph{iterative} manner, rather than by initializing $\left\{\mu_{j}\right\}_{j=1}^{C}$ once and then separating different clusters based only on initialization results. Additionally, contrary to us, the conventional methods for dealing with inter-manifold separation typically impose push-away constraints on all data points from different manifolds \cite{zhang2018semi,yang2016multi}, defined as:
\vspace{-1em}
\begin{equation}
    L_{sep}=-\sum_{i=1}^{N} \sum_{j=1}^{N}d_Z\left(z_i, z_j\right) \cdot \pi(l(x_i) \ne l(x_j))
\end{equation}
\indent The main differences between $L_{rank}$ and $L_{sep}$ are as follows: (1) $L_{sep}$ imposes constraints on embedding $\left\{z_{i}\right\}_{i=1}^{N}$, which indirectly affects the network parameters $\theta_f$. In contrast,  $L_{rank}$ imposes constrains directly on parameters $\left\{\mu_{j}\right\}_{j=1}^{C}$ in the form of \emph{regularization} item. (2) $L_{rank}$ is easier to optimize, faster to process, and more accurate. $L_{sep}$ is imposed on all data points from different manifolds, which involves $N$$\times$$N$ \emph{point-to-point} relationships. This means that each point may be subject to the push-away force from other manifolds, but at the same time, each point has to meet the isometry constraint with its neighboring points. Under these two constraints, optimization is difficult and it is easy to fall into a local optimal solution. In contrast, $L_{rank}$ is imposed directly on the clustering centers, involving only $C$$\times$$C$ \emph{cluster-to-cluster} relationships, which avoids the above problem and makes it easier to optimize. (3) The parameter $\kappa$ introduced in $L_{rank}$ allows us to control the extent of separation between manifolds.

\textbf{Alignment Loss.}
Global ranking loss $L_{rank}$ is imposed directly on $\left\{\mu_{j}\right\}_{j=1}^{C}$, so optimizing $L_{rank}$ only updates $\left\{\mu_{j}\right\}_{j=1}^{C}$ rather the encoder's parameter $\theta_f$. However, the optimization of $\left\{\mu_{j}\right\}_{j=1}^{C}$ not only relies on $L_{rank}$, but is also constrained by $L_{cluster}$, which ensures that data points remain roughly distributed around cluster centers and do not deviate significantly from them during the optimization process. Alignment loss $L_{align}$, as an auxiliary term, aims to help align learnable centers $\left\{\mu_{j}\right\}_{j=1}^{C}$ with the ground-truth centers $\{v^Z_j\}_{j=1}^{C}$ and \emph{make this binding stronger}, defined as
\vspace{-1em}
\begin{equation}
    L_{align}=\sum_{j=1}^C ||\mu_j-v^Z_{j}||
\end{equation}

\vspace{-0.5em}
\noindent where $\{v^Z_j\}_{j=1}^{C}$ are defined as $v^Z_j=\frac{1}{|M_j|}\sum_{i\in M_j}z_i$. The derivation for the gradient of $L_{align}$ with respect to center $\mu_{j}$ is placed in \textbf{Appendix A.2}. As shown in Fig.~\ref{fig:1}, three structure-oriented losses $\{L_{rank}(x, \mu),L_{\textit{LIS}(x, z)},L_{align}(z, \mu)\}$, form a closed loop between input $x$, embeddings $z$, and cluster centers $\mu$. 

\vspace{-0.5em}
\subsection{Training Strategy}\label{sec:3.3}
\begin{figure*}[!htbp]
	\begin{center}
		\subfigure[]{\includegraphics[width=0.11\linewidth]{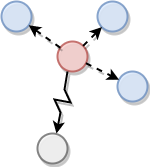}}
		\subfigure[Schematic of training strategy]{\includegraphics[width=0.63\linewidth]{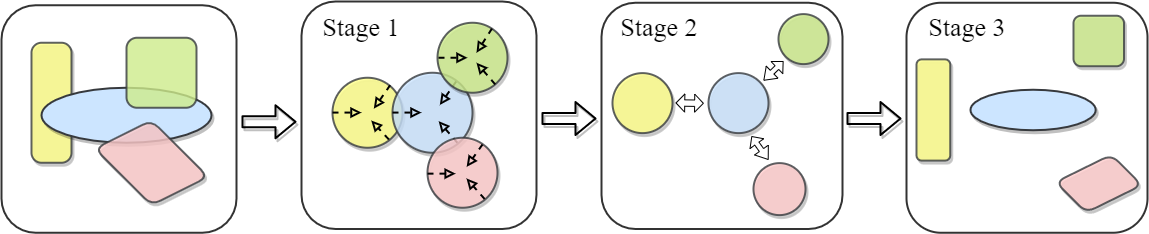}}
		\subfigure[Learning curve]{\includegraphics[width=0.24\linewidth]{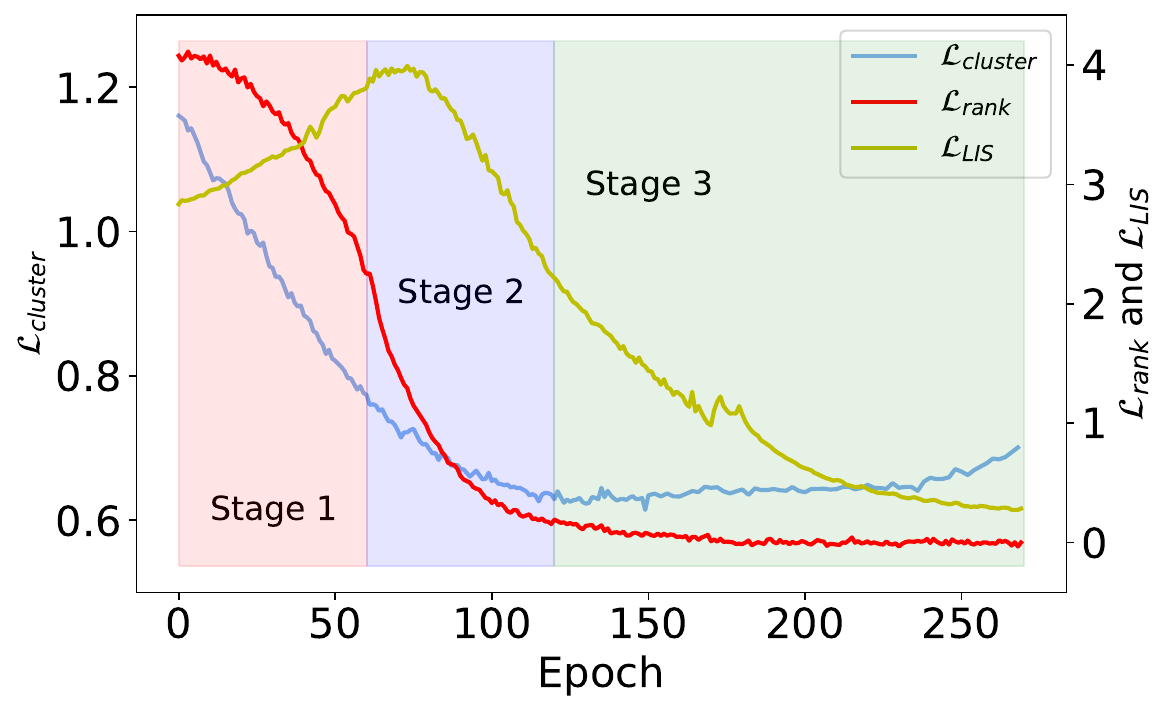}}
	\end{center}
	\vspace{-1em}
	\caption{Fig.~\ref{fig:2} (a) is the force analysis of the contradiction between clustering and local structure preservation. Fig.~\ref{fig:2} (b) is the schematic of training strategy, where four different colors and shapes represent four intersecting manifolds, and three stages involve the manifold clustering, separation, and geometric recovery. Fig.~\ref{fig:2} (c) is the learning curve of losses $L_{cluster}$, $L_{rank}$, $L_{\textit{LIS}}$ on the MNIST-test dataset.}
	\label{fig:2}
\end{figure*}

\vspace{-0.5em}
\textbf{Contradiction.} 
The contradiction between clustering and local structure preservation is analyzed from the \emph{forces analysis} perspective. As shown in Fig.~\ref{fig:2} (a), we assume that there exists a data point ({\color[rgb]{1,0,0}red} point) and its three nearest neighbors ({\color[rgb]{0,0,1}blue} points) around a cluster center ({\color[rgb]{0.7,0.7,0.7}gray} point). When clustering and local structure preserving are optimized simultaneously, it's easy to fall into a local optimum, where the data point is in steady-state, and the resultant force from its three nearest neighbors is equal in magnitude and opposite to the gravitational forces of the cluster.

\textbf{Alternating Training.} 
To solve the above problem and integrate the goals of clustering and local structure preservation into a unified framework, we take an alternating training strategy. Within each epoch, we first optimize $L_{cluster}$ and $L_{rank}$ in a \emph{mini-batch}, with joint loss defined as 
\vspace{-0.5em}
\begin{equation}
    L_{1}=\alpha L_{cluster}+L_{rank}
\end{equation}
Then at each epoch, we jointly optimize isometry loss $L_{\textit{LIS}}$ and $L_{align}$ on \emph{the whole dataset}, defined as 
\vspace{-0.5em}
\begin{equation}
    L_2=\beta L_{\textit{LIS}}+L_{align}
\end{equation}

\vspace{-0.5em}
\textbf{Weight Continuation.} 
At different stages of training, we have different requirements for clustering and structure preservation. At the beginning of training, to successfully decouple the overlapping manifolds, we hope that the $L_{cluster}$ will dominate and $L_{\textit{LIS}}$ will be auxiliary. When the margin between different manifolds is sufficiently large, the weight $\alpha$ for $L_{cluster}$ can be gradually reduced, while the weight $\beta$ for $L_{\textit{LIS}}$ can be gradually increased, focusing on the preservation of the local isometry. The whole algorithm is summarized in Algorithm 1.

\vspace{-0.5em}
\begin{algorithm}[H]
\footnotesize
	\caption{Algorithm for GCML}
	\label{algo:1}
	\begin{algorithmic}[1]
		\Require Input samples: $X$; Number of clusters: $C$; Number of batches: $B$; Number of iterations: $E$. 
		
		\Ensure Autoencoder's weights: $\theta_f$ and $\theta_g$; Cluster labels $\left\{s_{i}\right\}_{i=1}^{N}$; Trainable cluster centers $\left\{\mu_{j}\right\}_{j=1}^{C}$.
		
		\State Initialize the weight $\left\{\mu_{j}\right\}_{j=1}^{C}$, $\theta_f$ and $\theta_g$, and obtain initialized soft label assignment $\left\{s_{i}\right\}_{i=1}^{N}$.
		\For{$epoch$ $\in$ \{0,1,$\cdots$,$E$-1\}}
		\State Compute embedded points $\left\{z_{i}\right\}_{i=1}^{N}$ and distribution $Q$; 
		\State Update target distribution $P$;
		\State Compute soft cluster centers $\left\{v^X_i\right\}_{i=1}^{C}$ and $\left\{v^Z_i\right\}_{i=1}^{C}$.
		
		\For{$batch$ $\in$ \{0,1,$\cdots$,$B$\}}
		\State Pick up one batch of samples $X_{batch}$ from $X$;
		\State Compute corresponding distribution $Q_{batch}$ and
		\State it's reconstruction $Y_{batch}$; 
		\State Pick up target distribution batch $P_{batch}$ from $P$;
		\State Compute loss $L_{ae}$, $L_{cluster}$ and $L_{rank}$;
		\State Update the weight $\theta_f$, $\theta_g$ and $\left\{\mu_{j}\right\}_{j=1}^{C}$.
		\EndFor
		\State Compute $L_{iso}$ and $L_{align}$ on all samples;
		\State Update the weight $\theta_f$ and $\left\{\mu_{j}\right\}_{j=1}^{C}$;
		\State Assign new soft labels $\left\{s_{i}\right\}_{i=1}^{N}$.
		\EndFor
		\State return $\theta_f$, $\theta_g$, $\left\{s_{i}\right\}_{i=1}^{N}$, $\left\{\mu_{j}\right\}_{j=1}^{C}$.
	\end{algorithmic}
\end{algorithm}

\vspace{-1em}

\textbf{Three-stage Explanation.} 
The training process can be roughly divided into three stages, as shown in Fig.~\ref{fig:2} (b), to explain the training strategy more vividly. Also, we provide the learning curves of key losses $L_{cluster}$, $L_{rank}$, $L_{\textit{LIS}}$ on the MNIST-test dataset in Fig.~\ref{fig:2} (c). At first, four different manifolds overlap. At Stage 1, $L_{cluster}$ dominates, thus data points within each manifold converge towards cluster centers to form spheres, but the local structure of manifolds is corrupted. At Stage 2, $L_{rank}$ dominates, thus different manifolds in the latent space move away from each other to increase the manifold margin and enhance the discriminability. At stage 3, these manifolds gradually recover their original local structure from the spherical shape with $L_{\textit{LIS}}$ dominating. Note that the above losses may coexist rather than being completely independent at different stages, but the role played by different losses varies due to the alternating training and weight continuation strategies.

\begin{table*}[!htbp]
	\begin{center}
	\caption{Clustering performance (ACC/NMI) of different algorithms on seven datasets. The best metrics are marked in \textbf{bold}.}
	\label{tab:1}
	\resizebox{\textwidth}{!}{
	\begin{tabular}{lcccccccc}
		\hline
		\multicolumn{1}{l}{\textbf{Algorithms}} &
		\multicolumn{1}{c}{\textbf{MNIST-full}} &
		\multicolumn{1}{c}{\textbf{MNIST-test}} &
		\multicolumn{1}{c}{\textbf{USPS}} &
		\multicolumn{1}{c}{\textbf{Fashion-MNIST}} &
		\multicolumn{1}{c}{\textbf{REUTERS-10K}} &
		\multicolumn{1}{c}{\textbf{HAR}} & 
		\multicolumn{1}{c}{\textbf{pendigits}} & \\ \hline
		$K$-Means \cite{bishop2006pattern}    & 0.532/0.500          & 0.546/0.501          & 0.668/0.601          & 0.474/0.512          & 0.599/0.375*    & 0.599/0.588  & 0.666/0.681     \\
		SC-Ncut \cite{bishop2006pattern}    & 0.656/0.731          & 0.660/0.704          & 0.649/0.794          & 0.508/0.575          & 0.658/0.401*   &0.538/0.741   & 0.724/0.784   \\
		GMM \cite{bishop2006pattern}    & 0.389/0.333          & 0.464/0.465          & 0.562/0.540          & 0.463/0.514          & 0.613/0.388*   & 0.585/0.648  & 0.673/0.682    \\
		AE+$K$-Means \cite{tian2014learning}& 0.818/0.747          & 0.815/0.784*         & 0.662/0.693          & 0.566/0.585*          & 0.721/0.432*   & 0.674/0.670*  & 0.713/0.733*  \\
		DEC   \cite{xie2016unsupervised}     & 0.903/0.854*          & 0.885/0.851*          & 0.889/0.873*          & 0.554/0.576*          & 0.773/0.528*   & 0.759/0.695*  & 0.679/0.671*    \\
		IDEC    \cite{guo2017improved}   & 0.918/0.868*          & 0.876/0.817*          & 0.893/0.876*          & 0.572/0.601*          & 0.785/0.541*    &0.786/0.718*   & 0.739/0.757*  \\
		VaDE \cite{jiang2016variational}    & 0.945/0.876          & 0.287/0.287          & 0.566/0.512          & 0.578/0.630          & 0.795/0.556*   & 0.801/0.720*  & 0.762/0.743*  \\
		DEPICT \cite{ghasedi2017deep}    & 0.965/0.917          & 0.963/0.915          & 0.899/0.906          & 0.392/0.392          & -   & -   & -   \\
		JULE  \cite{yang2016joint}     & 0.964/0.913          & 0.961/0.915          & 0.950/0.913          & 0.563/0.608          & -    & -   & -  \\
		DSC   \cite{yang2019deep}     & 0.978/0.941          & \textbf{0.980/0.946} & 0.869/0.857          & 0.662/0.645          & -    & -   & -  \\
		{\color[rgb]{0.75,0.75,0.75} ASPC-DA \cite{guo2019adaptive}}   & {\color[rgb]{0.75,0.75,0.75} 0.988/0.966}          & {\color[rgb]{0.75,0.75,0.75} 0.973/0.936}          & {\color[rgb]{0.75,0.75,0.75} 0.982/0.951}          & {\color[rgb]{0.75,0.75,0.75} 0.591/0.654}          & {\color[rgb]{0.75,0.75,0.75} -}  & {\color[rgb]{0.75,0.75,0.75} -}      & {\color[rgb]{0.75,0.75,0.75} -}              \\
		ASPC (w/o DA) \cite{guo2019adaptive} & 0.931/0.886*     & 0.813/0.792*        & 0.768/0.803*        & 0.576/0.632*        & 0.692/0.418*   & 0.769/0.682* & 0.769/0.751* \\
		N2D   \cite{mcconville2019n2d}     & 0.969/0.928*          & 0.954/0.897*          & 0.954/0.898*          & 0.671/0.678*          & 0.784/0.536*    & 0.807/0.721*   & 0.847/0.808*   \\
		GCML (ours) & \textbf{0.980/0.946} & 0.972/0.930          & \textbf{0.958/0.902} & \textbf{0.710/0.685} & \textbf{0.836/0.590} & \textbf{0.844/0.762} & \textbf{0.855/0.814} \\ \hline
	\end{tabular}}
	\end{center}
\end{table*}

\section{Experiments}
\subsection{Experimental Setups} \label{sec:4.1}
In this section, the effectiveness of GCML is evaluated in 7 benchmark datasets: MNIST-full, MNIST-test, USPS, Fashion-MNIST, REUTERS-10K, HAR, and Pendigits on which GCML is compared with 12 other methods in 8 evaluation metrics including metrics specifically designed for clustering and multi-manifold learning. The brief descriptions of the datasets are given in \textbf{Appendix A.1}.

\textbf{Evaluation Metrics.}
Two standard evaluation metrics: Accuracy (ACC) and Normalized Mutual Information (NMI) are used to evaluate clustering performance. Besides, six evaluation metrics are adopted in this paper to evaluate the performance of multi-manifold learning, including Relative Rank Error (RRE), Trustworthiness (Trust), Continuity (Cont), Root Mean Reconstruction Error (RMRE), Locally Geometric Distortion (LGD) and Cluster Rank Accuracy (CRA). Limited by space, their precise definitions are available in \textbf{Appendix A.3}.

\textbf{Parameters Settings.}
The encoder structure is $d$-500-500-500-2000-10 where $d$ is the dimension of the input data, and the decoder is its mirror. After pretraining, to initialize the learnable clustering centers, the t-SNE is applied to find the best clustable manifold in the latent space $Z$, and then the $K$-Means algorithm is run to obtain the label assignments for each data point. The centers of each category in the latent space $Z$ are set as initial cluster centers $\left\{\mu_{j}\right\}_{j=1}^{C}$. Besides, Adam optimizer with learning rate $\lambda$=0.001 is used, the batch size is set to 256, the epoch is set to 300, the parameter $k$ for nearest neighbor is set to 5, and the parameter $\kappa$ is set to 3 for all datasets. Sensitivity analysis for parameters $k$ and $\kappa$ is available in \textbf{Appendix A.4}.  As described in Sec.~\ref{sec:3.3}, the weight continuation is applied to train the model. The weight parameter $\alpha$ for $L_{cluster}$ decreases linearly from 0.1 to 0 within epoch 0-150. In contrast, the weight parameter $\beta$ for $L_{\textit{LIS}}$ increases linearly from 0 to 1.0 within epoch 0-150. In this paper, each set of experiments is run 5 times with different random seeds, and the results are averaged into the final performance metrics.

\begin{figure*}[!htbp]
\small
	\begin{center}
		\includegraphics[width=1.0\linewidth]{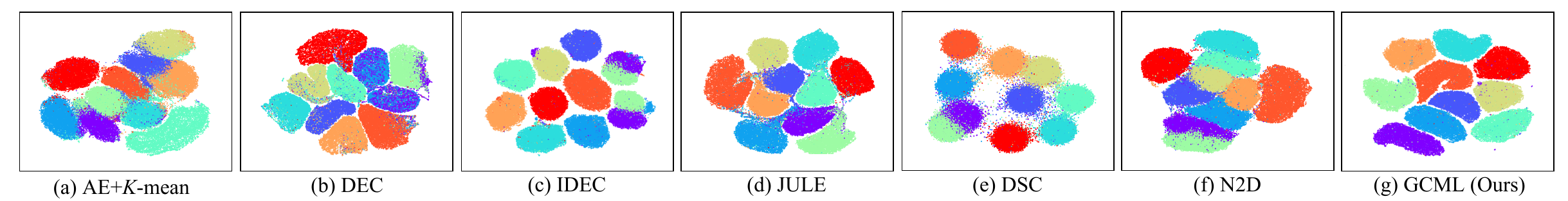}
	\end{center}
	\vspace{-1em}
	\caption{Visualization of the embeddings learned by different algorithms on the MNIST-full dataset.}
	\label{fig:3}
	\vspace{-0.5em}
\end{figure*}

\begin{figure*}[!htbp]
	\begin{center}
        \includegraphics[width=0.76\linewidth]{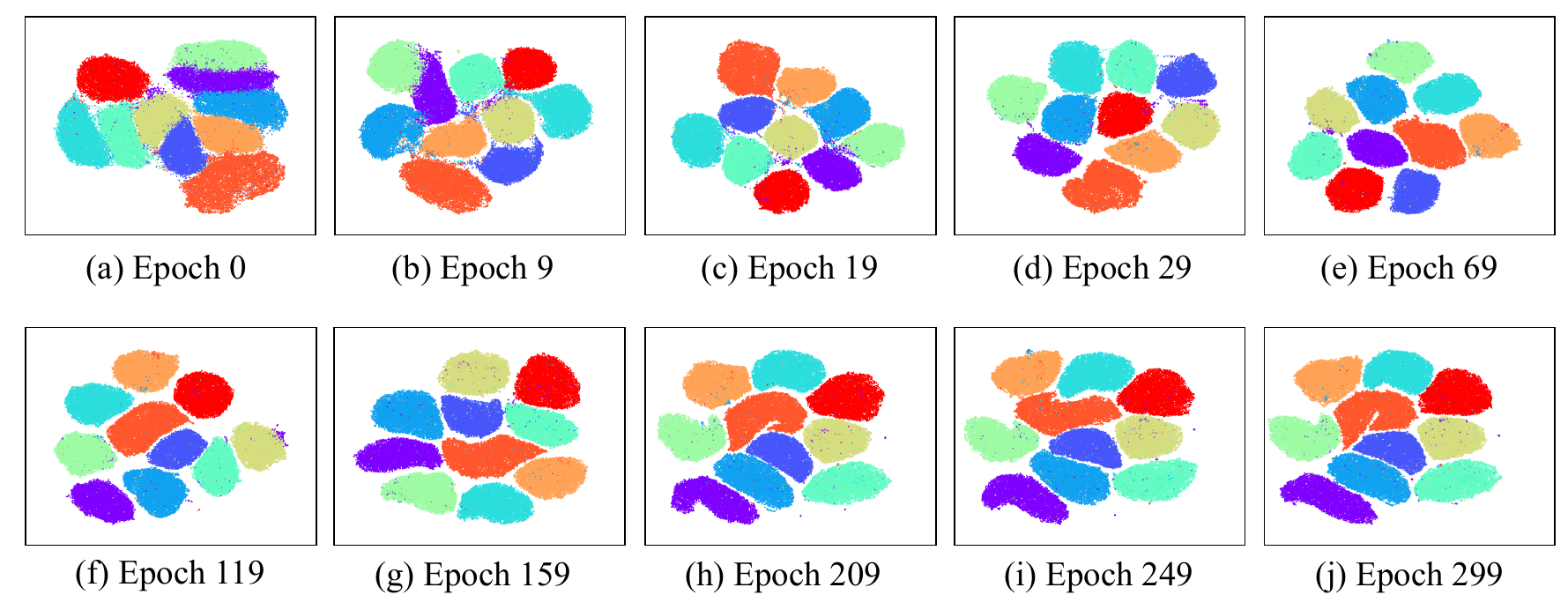}
	\end{center}
	\vspace{-1em}
	\caption{Clustering visualization at different stages during the training process on the MNIST-full dateset.}
	\label{fig:4}
\end{figure*}

\subsection{Evaluation of Clustering}
\textbf{Quantitative Comparison.}
The metrics ACC/NMI of different methods are reported in Tab.~\ref{tab:1}. For those methods whose results are not reported or experimental settings are not clear, we run the released code with the same provided hyperparameters and mark them with (*). Moreover, we mark those methods that are only applicable to 2-D image data as (-) on the \emph{vector} dataset. While ASPC-DA achieves the best performance on three datasets (MNIST-test, MNIST-full, and USPS), its performance gains do not come directly from clustering, but from sophisticated modules such as data augmentation and self-paced learning. Once these modules are removed, there is large performance degradation. For example, with Data Augmentation (DA) removed, ASPC achieves less competitive performance, e.g., an accuracy of 0.931 (\textit{vs} 0.988) on MNIST-full, 0.813 (\textit{vs} 0.973) on MNIST-test and 0.768 (\textit{vs} 0.982) on USPS. Since ASPC-DA is based on the MLP architecture, its image-based augmentation cannot be applied directly to vector data, which explains why ASPC has no performance advantage (even compared to DEC and IDEC) on the vector datasets, such as REUTERS-10K and HAR datasets.

In a fair comparison (without considering ASPC-DA and marking its results in Tab.~\ref{tab:1} as {\color[rgb]{0.7,0.7,0.7}gray} color), we find that GCML outperforms $K$-Means, GNN, and SC-Ncut by a significant margin and surpasses the other nine compared DNN-based algorithms on all datasets except MNIST-test. Nevertheless, even with the MNIST-test dataset, GCML still ranks second, outperforming the third by 0.9\%. In particular, we obtain the best performance on the Fashion-MNIST, REUTERS-10K, and HAR datasets, and more notably, our clustering accuracy exceeds the current SOTA method by 3.9\%, 4.1\%, and 3.8\%, respectively.

\textbf{Clustering Visualization.} 
The visualization of GCML with comparison methods is shown in Fig.~\ref{fig:3} (visualized using UMAP). Among all methods, only DEC, IDEC, and GCML can hold clear boundaries between different clusters, while the cluster boundaries of the other methods are indistinguishable. Though DEC and IDEC successfully separate different clusters, they group many data points from different classes into the same cluster. Most importantly, due to the use of the clustering-oriented loss, the embedding learned by these algorithms (especially DSC) \emph{tend to form spheres and disrupt the original topological structure}. Instead, GCML overcomes the above problems and achieves almost perfect separation between different clusters while preserving the local and global structure. 

\begin{figure*}[ht]
	\begin{center}
		\includegraphics[width=1.0\linewidth]{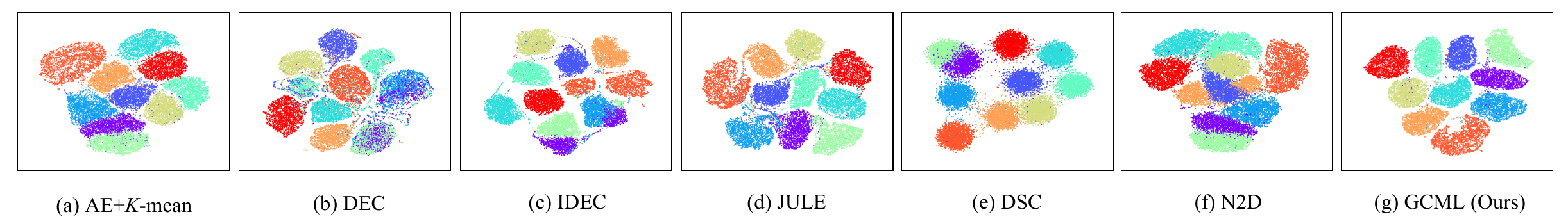}
	\end{center}
	\caption{Visualization of the embeddings learned from testing samples on the MNIST-full dataset.}
	\label{fig:5}
\end{figure*}

\begin{figure*}[!htbp]
	\begin{center}
		\includegraphics[width=1.0\linewidth]{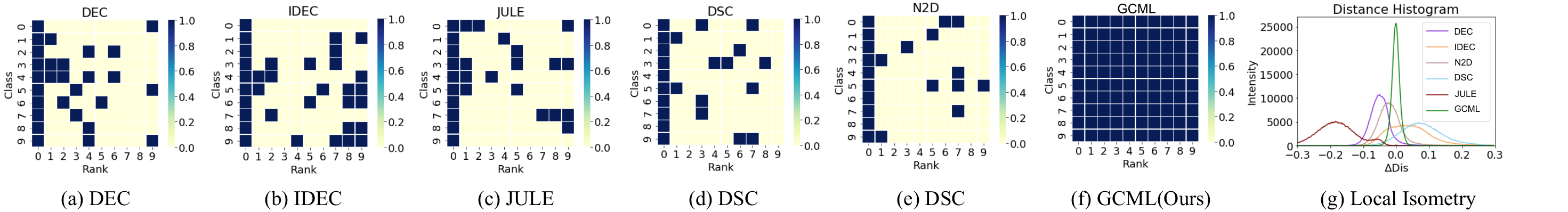}
	\end{center}
	\caption{Statistical analysis for evaluating the capability of geometric structure preservation from the input space to the latent space.}
	\label{fig:6}
\end{figure*}

The embeddings of the latent space during the training process are visualized in Fig.~\ref{fig:4} for explaining how both clustering and structure-preserving are achieved. We can see that the different clusters initialized by the pre-trained autoencoder are closely adjacent. In the early stage of training, with clustering loss $L_{cluster}$ and global ranking loss $L_{rank}$, different manifolds are separated from each other, but each manifold loses its local structure, and all of them degenerate into spheres. As the training progresses, the weight $\alpha$ for $L_{cluster}$ gradually decreases, while the weight $\beta$ for $ L_{\textit{LIS}}$ increases and \textit{the optimization is gradually focused from global to local}, with each manifold gradually recovering its original geometric structure from the sphere. These visualizations show that clustering-oriented loss does deteriorate the geometric structure of the latent space, and the designed structure-oriented losses help to recover it.

\textbf{Generalizability Evaluation.}
Tab.~\ref{tab:2} demonstrates that a learned GCML can generalize well to unseen data with high clustering accuracy. Taking MNIST-full as an example, GCML was trained using 50,000 training samples and then tested on the remaining 20,000 testing samples using the learned model. In terms of the metrics ACC and MNI, GCML is optimal for both training and testing samples. More importantly, there is hardly any degradation in the performance of GCML on the testing samples compared to the training samples, while all other methods show a significant drop in performance, e.g., DEC from 84.1\% to 74.8\%. This demonstrates the importance of geometric structure preservation for good generalizability. The visualization results of the testing samples are shown in Fig.~\ref{fig:5}; even for testing samples, GCML still shows distinguishable inter-cluster discriminability, while other methods couple different clusters together, which shows GCML's great generalizability.

\begin{table}[!htbp]
\begin{center}
\caption{Generalizability evaluated by ACC/NMI.}
\label{tab:2}
	\begin{tabular}{lcc}
		\hline
		\multicolumn{1}{l}{\textbf{Algorithms}} & \multicolumn{1}{c}{\textbf{training samples}} & \multicolumn{1}{c}{\textbf{testing samples}} \\ \hline
		AE+$K$-Means & 0.815/0.736          & 0.751/0.711        \\
		DEC        & 0.841/0.773          & 0.748/0.704          \\
		IDEC       & 0.845/0.860          & 0.826/0.842          \\
		JULE       & 0.958/0.907          & 0.921/0.895          \\
		DSC        & 0.975/0.939          & 0.969/0.921          \\
		N2D        & 0.974/0.930          & 0.965/0.911          \\
		GCML (ours) & \textbf{0.978/0.941} & \textbf{0.978/0.941}\\ \hline
	\end{tabular}
\end{center}
\end{table}
\begin{table*}[!htbp]
	\begin{center}
	\caption{Performance for multi-manifold learning (left) and downstream tasks (right) on the MNIST-full and Fashion-MNIST datasets.}
	\label{tab:3}
	\resizebox{\textwidth}{!}{
	\begin{tabular}{llcccccc|cccc}
		\hline
		\textbf{Datasets} &
		\multicolumn{1}{l}{\textbf{Algorithms}} &
		\multicolumn{1}{l}{\textbf{RRE}$\downarrow$} &
		\multicolumn{1}{c}{\textbf{Trust}$\uparrow$} &
		\multicolumn{1}{c}{\textbf{Cont}$\uparrow$} &
		\multicolumn{1}{c}{\textbf{$d$-RMSE}$\downarrow$} &
		\multicolumn{1}{c}{\textbf{LGD}$\downarrow$} &
		\multicolumn{1}{c}{\textbf{CRA}$\uparrow$} &
		\multicolumn{1}{c}{\textbf{MLP}$\uparrow$} &
		\multicolumn{1}{c}{\textbf{RFC}$\uparrow$} &
		\multicolumn{1}{c}{\textbf{SVM}$\uparrow$} &
		\multicolumn{1}{c}{\textbf{LR}$\uparrow$} \\ \hline
		\multirow{6}{*}{MNIST-full}   & DEC        & 0.09988          & 0.84499          & 0.94805          & 44.8535          & 4.37986          & 0.28   & 0.8647          & 0.8706          & 0.8707          & 0.8566      \\
		& IDEC       & 0.00984          & 0.99821          & 0.97936          & 24.5803          & 1.71484          & 0.33   & 0.9797          & 0.9737          & 0.9852          & 0.9650      \\
		& JULE       & 0.02657          & 0.93675          & 0.98321          & 28.3412          & 2.12955          & 0.27   & 0.9802          & 0.9825          & 0.9787          & 0.9743      \\
		& DSC        & 0.09785          & 0.87315          & 0.92508          & 6.98098          & 1.19886          & 0.23   & 0.9622          & 0.9501          & 0.9837          & 0.9752      \\
		& N2D        & 0.01002          & 0.99243          & 0.98466          & 5.7162           & 0.69946          & 0.21   & 0.9796          & 0.9803          & 0.9799          & 0.9792      \\
		& GCML (ours) & \textbf{0.00567} & \textbf{0.99978} & \textbf{0.98716} & \textbf{5.4986}  & \textbf{0.69168} & \textbf{1.00} & \textbf{0.9851} & \textbf{0.9874} & \textbf{0.9869} & \textbf{0.9841}\\ \hline
		
		\multirow{6}{*}{Fashion-MNIST} & DEC        & 0.04787          & 0.93896          & 0.95450          & 39.3274          & 3.87731          & 0.37   & 0.6268          & 0.9853          & 0.6377          & 0.6245      \\
		& IDEC       & 0.01089          & 0.99683          & 0.97797          & 25.4024          & 1.91385          & 0.27      & 0.8367          & 0.9918          & \textbf{0.8607} & 0.7514   \\
		& JULE       & 0.03013          & 0.97732          & 0.97923          & 15.2213          & 1.43642          & 0.43      & 0.8541          & 0.9892          & 0.8566          & 0.7723   \\
		& DSC        & 0.05168          & 0.95013          & 0.96121          & 17.2201          & 1.42091          & 0.36      & 0.8084          & 0.9823          & 0.8618          & 0.7676   \\
		& N2D        & 0.00894          & 0.99062          & 0.98054          & 14.49079         & \textbf{1.28180} & 0.26      & 0.8412          & 0.9493          & 0.8230          & 0.7753   \\
		& GCML (ours) & \textbf{0.00836} & \textbf{0.99868} & \textbf{0.98203} & \textbf{13.3788} & 1.33893          & \textbf{1.00} & \textbf{0.8642} & \textbf{0.9942} & 0.8468          & \textbf{0.7768}\\ \hline
	\end{tabular}}
	\end{center}
\end{table*}
\begin{table*}[!htbp]
\begin{center}
\caption{Ablation study of loss items and training strategies used in the proposed GCML framework.}
\label{tab:4}
\begin{tabular}{cllllllll}
\hline
\textbf{Datasets} &
  \multicolumn{1}{c}{\textbf{Methods}} &
  \multicolumn{1}{c}{\textbf{ACC/NMI}$\uparrow$} &
  \multicolumn{1}{c}{\textbf{RRE}$\downarrow$} &
  \multicolumn{1}{c}{\textbf{Trust}$\uparrow$} &
  \multicolumn{1}{c}{\textbf{Cont}$\uparrow$} &
  \textbf{$d$-RMSE}$\downarrow$ &
  \textbf{LGD}$\downarrow$ &
  \textbf{CRA}$\uparrow$ \\ \hline
 &
  w/o SL &
  0.976/0.939 &
  0.0093 &
  0.9967 &
  0.9816 &
  24.589 &
  1.6747 &
  0.32 \\
 &
  {\color[rgb]{0.75,0.75,0.75} w/o CL} &
  {\color[rgb]{0.75,0.75,0.75} 0.814/0.736} &
  {\color[rgb]{0.75,0.75,0.75} 0.0004} &
  {\color[rgb]{0.75,0.75,0.75} 0.9998} &
  {\color[rgb]{0.75,0.75,0.75} 0.9990} &
  {\color[rgb]{0.75,0.75,0.75} 7.458} &
  {\color[rgb]{0.75,0.75,0.75} 0.0487} &
  {\color[rgb]{0.75,0.75,0.75} 1.00} \\
 &
  w/o WC &
  0.977/0.943 &
  0.0065 &
  0.9987 &
  0.9860 &
  5.576 &
  0.6968 &
  0.98 \\
 &
  w/o AT &
  0.978/0.944 &
  0.0069 &
  0.9986 &
  0.9851 &
  5.617 &
  0.7037 &
  0.96 \\
\multirow{-5}{*}{MNIST-full} &
  full model &
  \textbf{0.980/0.946} &
  \textbf{0.0056} &
  \textbf{0.9997} &
  \textbf{0.9871} &
  \textbf{5.498} &
  \textbf{0.6916} &
  \textbf{1.00} \\ \hline
 &
  w/o SL &
  0.706/0.682 &
  0.0108 &
  0.9964 &
  0.9781 &
  25.954 &
  1.8936 &
  0.30 \\
 &
  {\color[rgb]{0.75,0.75,0.75} w/o CL} &
  {\color[rgb]{0.75,0.75,0.75} 0.576/0.569} &
  {\color[rgb]{0.75,0.75,0.75} 0.0004} &
  {\color[rgb]{0.75,0.75,0.75} 0.9994} &
  {\color[rgb]{0.75,0.75,0.75} 0.9995} &
  {\color[rgb]{0.75,0.75,0.75} 7.654} &
  {\color[rgb]{0.75,0.75,0.75} 0.0523} &
  {\color[rgb]{0.75,0.75,0.75} 1.00} \\
 &
  w/o WC &
  0.702/0.695 &
  0.0084 &
  0.9972 &
  0.9814 &
  \textbf{13.238} &
  1.3474 &
  \textbf{1.00} \\
 &
  w/o AT &
  0.708/0.694 &
  0.0097 &
  0.9975 &
  0.9798 &
  13.354 &
  1.3611 &
  \textbf{1.00} \\

\multirow{-5}{*}{Fashion-MNIST} &
  full model &
  \textbf{0.710/0.685} &
  \textbf{0.0083} &
  \textbf{0.9986} &
  \textbf{0.9820} &
  13.378 &
  \textbf{1.3389} &
  \textbf{1.00} \\ \hline

\end{tabular}
\end{center}
\end{table*}

\vspace{-1em}
\subsection{Evaluation of Multi-Manifold Learning}
\textbf{Quantitative Metrics.}
Though many previous works have claimed that they brought clustering and dimensionality reduction into a unified framework, unfortunately, they all lacked an analysis of the effectiveness of the learned embeddings. In this paper, we compare GCML with the other five methods in six quantitative metrics on seven datasets. Limited by space, only the results of MNIST-full and Fashion-MNIST are provided on the left side of Tab.~\ref{tab:3} and more results are in \textbf{Appendix A.5}. The results show that GCML outperforms all other methods, especially in the CRA metric, which is not only the best on all datasets but reaches 1.0, which means that the ``rank" between different manifolds in the latent space is completely preserved and proves the effectiveness of the global ranking loss $L_{rank}$.

\textbf{Statistical Analysis.}
The statistical analysis is performed to show the extent to which local and global structure is preserved in the latent space for each algorithm. Taking MNIST-full as an example, the statistical analysis of the global rank-preservation is shown in Fig.~\ref{fig:6} (a)-(f). For the $i$-th cluster, if the rank (in terms of Euclidean distance) between it and the $j$-th cluster is preserved from the input space to the latent space, then the grid in the $i$-th row and $j$-th column is marked as blue, otherwise yellow. As shown in the figure, only GCML can fully preserve the global rank between different clusters, while all other methods fail. 

Moreover, we perform statistical analysis for the local isometry property of each algorithm. For each sample $x_i$, it forms a number of point pairs with its neighborhood samples $\{(x_i, x_j)|i=1,2,...,N;x_j \in \mathcal{N}_{i}^{X}\}$. We compute the differences in the distance of these point pairs from the input space to the latent space $\{d_Z(x_i, x_j)-d_X(x_i, x_j)|i=1,2,...,N;x_j \in \mathcal{N}_{i}\}$, and plot them as a histogram. As shown in Fig.~\ref{fig:6} (g),  the curve of GCML are distributed on both sides of the 0 value, with \emph{maximum peak height} and \emph{minimum peak-bottom width}, respectively, which indicates that GCML achieves the best local isometry. Though IDEC \cite{guo2017improved} claims that they can preserve the local structure well, their results are still far from ours.

\textbf{Downstream Tasks.}
Numerous deep clustering algorithms have recently claimed to obtain meaningful low-dimensional embeddings; however, they have not analyzed and experimented with the ``meaningful" ones. Therefore, we are interested in whether these proposed methods can really learn manifold embeddings that are useful for downstream tasks. Four different classifiers, including a linear classifier (Logistic Regression; LR), two nonlinear classifiers (MLP, SVM), and a tree-based classifier (Random Forest Classifier; RFC) are used as downstream tasks, all of which use default parameters and default implementations in sklearn \cite{scikit-learn} for a fair comparison. The learned embeddings are \emph{frozen} and used as input for training. The classification accuracy evaluated on the test set serves as a metric to evaluate the effectiveness of learned embeddings. Limited by space, only the results of MNIST-full and Fashion-MNIST are provided on the right side of Tab.~\ref{tab:3} and more results are in \textbf{Appendix A.5}. The results show that GCML outperforms the other methods overall on all seven datasets, with MLP, RFC, and LR as downstream tasks.

\subsection{Ablation Study}
Tab.~\ref{tab:4} evaluates the effectiveness of the proposed loss terms and training strategies with 5 sets of experiments: the model without (A) Structure-oriented Loss (SL); (B) Clustering-oriented Loss (CL); (C) Weight Continuation (WC); (D) Alternating Training (AT), and (E) the full model. Limited by space, only the results MNIST-full and Fashion-MNIST are provided and more results are in \textbf{Appendix A.6}. After analyzing the results, we can conclude: (1) CL is the most important factor for obtaining excellent clustering performance, the lack of which leads to unsuccessful clustering, hence the numbers in the table are not meaningful and marked in {\color[rgb]{0.7,0.7,0.7}gray} color. (2) SL not only brings subtle improvements in clustering performance but greatly improves the performance of multi-manifold learning. (3) The training strategies (WC and AT) both improve the performance of clustering and multi-manifold learning to some extent, especially on metrics such as RRE, Trust, CRA, etc.

\vspace{-0.5em}
\section{Conclusion}
The proposed GCML framework imposes clustering-oriented and structure-oriented constraints to optimize the latent space for simultaneously performing clustering and multi-manifold learning with geometric structure preservation. Extensive experiments demonstrate that GCML is not only comparable to the SOTA clustering algorithms but learns effective manifold embeddings, which is beyond the capability of those algorithms that only care about clustering accuracy. Finally, GCML is appliable to generalized data with various dimensions, not limited to 2-D image data.

{\small
\bibliographystyle{ieee_fullname}
\bibliography{egbib}

\begin{thebibliography}{10}\itemsep=-1pt

\bibitem{bishop2006pattern}
Christopher~M Bishop.
\newblock {\em Pattern recognition and machine learning}.
\newblock springer, 2006.

\bibitem{gao2020clustering}
Zhangyang Gao, Haitao Lin, Stan Li, et~al.
\newblock Clustering based on graph of density topology.
\newblock {\em arXiv preprint arXiv:2009.11612}, 2020.

\bibitem{ghasedi2017deep}
Kamran Ghasedi~Dizaji, Amirhossein Herandi, Cheng Deng, Weidong Cai, and Heng
  Huang.
\newblock Deep clustering via joint convolutional autoencoder embedding and
  relative entropy minimization.
\newblock In {\em Proceedings of the IEEE international conference on computer
  vision}, pages 5736--5745, 2017.

\bibitem{guo2017improved}
Xifeng Guo, Long Gao, Xinwang Liu, and Jianping Yin.
\newblock Improved deep embedded clustering with local structure preservation.
\newblock In {\em IJCAI}, pages 1753--1759, 2017.

\bibitem{guo2019adaptive}
Xifeng Guo, Xinwang Liu, En Zhu, Xinzhong Zhu, Miaomiao Li, Xin Xu, and
  Jianping Yin.
\newblock Adaptive self-paced deep clustering with data augmentation.
\newblock {\em IEEE Transactions on Knowledge and Data Engineering}, 2019.

\bibitem{he2020momentum}
Kaiming He, Haoqi Fan, Yuxin Wu, Saining Xie, and Ross Girshick.
\newblock Momentum contrast for unsupervised visual representation learning.
\newblock In {\em Proceedings of the IEEE/CVF Conference on Computer Vision and
  Pattern Recognition}, pages 9729--9738, 2020.

\bibitem{jiang2016variational}
Zhuxi Jiang, Yin Zheng, Huachun Tan, Bangsheng Tang, and Hanning Zhou.
\newblock Variational deep embedding: An unsupervised and generative approach
  to clustering.
\newblock {\em arXiv preprint arXiv:1611.05148}, 2016.

\bibitem{lecun1998gradient}
Yann LeCun, L{\'e}on Bottou, Yoshua Bengio, and Patrick Haffner.
\newblock Gradient-based learning applied to document recognition.
\newblock {\em Proceedings of the IEEE}, 86(11):2278--2324, 1998.

\bibitem{li2020consistent}
Stan~Z Li, Lirong Wu, and Zelin Zang.
\newblock Consistent representation learning for high dimensional data
  analysis.
\newblock {\em arXiv preprint arXiv:2012.00481}, 2020.

\bibitem{maaten2008visualizing}
Laurens van~der Maaten and Geoffrey Hinton.
\newblock Visualizing data using t-sne.
\newblock {\em Journal of machine learning research}, 9(Nov):2579--2605, 2008.

\bibitem{macqueen1965some}
J MacQueen.
\newblock Some methods for classification and analysis of multivariate
  observations.
\newblock In {\em Proc. 5th Berkeley Symposium on Math., Stat., and Prob}, page
  281, 1965.

\bibitem{mcconville2019n2d}
Ryan McConville, Raul Santos-Rodriguez, Robert~J Piechocki, and Ian Craddock.
\newblock N2d:(not too) deep clustering via clustering the local manifold of an
  autoencoded embedding.
\newblock {\em arXiv preprint arXiv:1908.05968}, 2019.

\bibitem{mcinnes2018umap}
Leland McInnes, John Healy, and James Melville.
\newblock Umap: Uniform manifold approximation and projection for dimension
  reduction.
\newblock {\em arXiv preprint arXiv:1802.03426}, 2018.

\bibitem{mukherjee2019clustergan}
Sudipto Mukherjee, Himanshu Asnani, Eugene Lin, and Sreeram Kannan.
\newblock Clustergan: Latent space clustering in generative adversarial
  networks.
\newblock In {\em Proceedings of the AAAI Conference on Artificial
  Intelligence}, volume~33, pages 4610--4617, 2019.

\bibitem{scikit-learn}
F. Pedregosa, G. Varoquaux, A. Gramfort, V. Michel, B. Thirion, O. Grisel, M.
  Blondel, P. Prettenhofer, R. Weiss, V. Dubourg, J. Vanderplas, A. Passos, D.
  Cournapeau, M. Brucher, M. Perrot, and E. Duchesnay.
\newblock Scikit-learn: Machine learning in {P}ython.
\newblock {\em Journal of Machine Learning Research}, 12:2825--2830, 2011.

\bibitem{ren2020deep}
Yazhou Ren, Ni Wang, Mingxia Li, and Zenglin Xu.
\newblock Deep density-based image clustering.
\newblock {\em Knowledge-Based Systems}, 197:105841, 2020.

\bibitem{roweis2000nonlinear}
Sam~T Roweis and Lawrence~K Saul.
\newblock Nonlinear dimensionality reduction by locally linear embedding.
\newblock {\em science}, 290(5500):2323--2326, 2000.

\bibitem{shaham2018spectralnet}
Uri Shaham, Kelly Stanton, Henry Li, Boaz Nadler, Ronen Basri, and Yuval
  Kluger.
\newblock Spectralnet: Spectral clustering using deep neural networks.
\newblock {\em arXiv preprint arXiv:1801.01587}, 2018.

\bibitem{shi2000normalized}
Jianbo Shi and Jitendra Malik.
\newblock Normalized cuts and image segmentation.
\newblock {\em IEEE Transactions on pattern analysis and machine intelligence},
  22(8):888--905, 2000.

\bibitem{tenenbaum2000global}
Joshua~B Tenenbaum, Vin De~Silva, and John~C Langford.
\newblock A global geometric framework for nonlinear dimensionality reduction.
\newblock {\em science}, 290(5500):2319--2323, 2000.

\bibitem{tian2014learning}
Fei Tian, Bin Gao, Qing Cui, Enhong Chen, and Tie-Yan Liu.
\newblock Learning deep representations for graph clustering.
\newblock In {\em Aaai}, volume~14, pages 1293--1299. Citeseer, 2014.

\bibitem{van2020scan}
Wouter Van~Gansbeke, Simon Vandenhende, Stamatios Georgoulis, Marc Proesmans,
  and Luc Van~Gool.
\newblock Scan: Learning to classify images without labels.
\newblock In {\em Proceedings of the European Conference on Computer Vision
  (ECCV)}, 2020.

\bibitem{xiao2017fashion}
Han Xiao, Kashif Rasul, and Roland Vollgraf.
\newblock Fashion-mnist: a novel image dataset for benchmarking machine
  learning algorithms.
\newblock {\em arXiv preprint arXiv:1708.07747}, 2017.

\bibitem{xie2016unsupervised}
Junyuan Xie, Ross Girshick, and Ali Farhadi.
\newblock Unsupervised deep embedding for clustering analysis.
\newblock In {\em International conference on machine learning}, pages
  478--487, 2016.

\bibitem{yang2016multi}
Bo Yang, Ming Xiang, and Yupei Zhang.
\newblock Multi-manifold discriminant isomap for visualization and
  classification.
\newblock {\em Pattern Recognition}, 55:215--230, 2016.

\bibitem{yang2016joint}
Jianwei Yang, Devi Parikh, and Dhruv Batra.
\newblock Joint unsupervised learning of deep representations and image
  clusters.
\newblock In {\em Proceedings of the IEEE Conference on Computer Vision and
  Pattern Recognition}, pages 5147--5156, 2016.

\bibitem{yang2019deep}
Xu Yang, Cheng Deng, Feng Zheng, Junchi Yan, and Wei Liu.
\newblock Deep spectral clustering using dual autoencoder network.
\newblock In {\em Proceedings of the IEEE Conference on Computer Vision and
  Pattern Recognition}, pages 4066--4075, 2019.

\bibitem{zhan2020online}
Xiaohang Zhan, Jiahao Xie, Ziwei Liu, Yew-Soon Ong, and Chen~Change Loy.
\newblock Online deep clustering for unsupervised representation learning.
\newblock In {\em Proceedings of the IEEE/CVF Conference on Computer Vision and
  Pattern Recognition}, pages 6688--6697, 2020.

\bibitem{zhang2018semi}
Yan Zhang, Zhao Zhang, Jie Qin, Li Zhang, Bing Li, and Fanzhang Li.
\newblock Semi-supervised local multi-manifold isomap by linear embedding for
  feature extraction.
\newblock {\em Pattern Recognition}, 76:662--678, 2018.

\end{thebibliography}
}

\clearpage
\appendix
\renewcommand\thefigure{A\arabic{figure}}
\renewcommand\thetable{A\arabic{table}}
\setcounter{table}{0}
\setcounter{figure}{0}

\section*{Appendix}

\section*{A.1 Datasets}
To show that our method works well with various kinds of datasets, we choose the following seven image and vector datasets. A brief description is given in Tab.~\ref{tab:A1}.

(1). MNIST-full \cite{lecun1998gradient}: The MNIST-full dataset consists of 70000 handwritten digits of 28 $\times$ 28 pixels. Each gray image is reshaped to a 784-dimensional vector.

(2). MNIST-test \cite{lecun1998gradient}: The MNIST-test is the testing part of MNIST dataset, which contains a total of 10000 samples.

(3). USPS: The USPS dataset is composed of 9298 gray-scale handwritten digit images with a size of 16$\times$16 pixels.
	
(4). Fashion-MNIST \cite{xiao2017fashion}: This Fashion-MNIST dataset has the same number of images and the same image size as MNIST-full, but it is fairly more complicated. Instead of digits, it consists of various types of fashion products.

(5). REUTERS-10K: REUTERS-10K is a subset of REUTERS with 10000 samples. Four root categories (corporate/industrial, government/social, markets, and economics) are used as labels, and the tf-idf features on the 2000 most frequent words are computed.
	
(6). HAR: HAR is a time-series dataset consisting of 10299 sensor samples from a smartphone. It was collected from 30 people performing six activities: walking, walking upstairs, walking downstairs, sitting, standing, and laying.

(7). Pendigits: HAR is a dataset consisting of 10992 samples from pressure-sensitive plates with ten different digits, each represented by 8 coordinates of the stylus.

\begin{table}[!htbp]
	\begin{center}
	\caption{Description of Datasets.}
	\label{tab:A1}
	\begin{tabular}{lccc}
		\hline
		\multicolumn{1}{l}{\textbf{Dataset}} & \multicolumn{1}{c}{\textbf{Samples}} & \multicolumn{1}{c}{\textbf{Categories}} & \multicolumn{1}{c}{\textbf{Data Size}} \\ \hline
		MNIST-full   & 70000 & 10 & 28$\times$28$\times$1 \\
		MNIST-test   & 10000 & 10 & 28$\times$28$\times$1 \\
		USPS         & 9298  & 10 & 16$\times$16$\times$1 \\
		Fashion-MNIST & 70000 & 10 & 28$\times$28$\times$1 \\
		REUTERS-10K  & 10000 & 4  & 2000    \\ 
		HAR  & 10299 & 6  & 561   \\ 
		Pendigits  & 10992 & 10  & 16   \\\hline
	\end{tabular}
	\end{center}
\end{table}

\section*{A.2 Gradient Derivation}
In the paper, we have emphasized time and again that $\left\{\mu_{j}\right\}_{j=1}^{C}$ is a set of \emph{\textbf{learnable}} parameters, which means that we can optimize it while optimizing the network parameter $\theta_f$. In Eq.~(4) of the paper, we have presented the gradient of $L_{cluster}$ with respect to $\mu_j$. In addition to $L_{cluster}$, both $L_{rank}$ and $L_{align}$ are involving $\mu_j$. Hence, the detailed derivations for the gradient of $L_{rank}$ and $L_{align}$ with respect to $\mu_j$ are also provided. The gradient of $L_{rank}$ with respect to each cluster center $\mu_{j}$ can be computed as:

\begin{equation*}
\begin{aligned}
\frac{\partial L_{rank}}{\partial \mu_j} 
& = \frac{\partial \sum_{i'=1}^{C} \sum_{j'=1}^{C}\left|d_Z\left(\mu_{i'}, \mu_{j'}\right)-\kappa*d_X\left(v^X_{i'}, v^X_{j'}\right)\right|}{\partial \mu_j}\\
& = \sum_{i'=1}^{C}\sum_{j'=1}^{C} \frac{\partial \left|d_Z\left(\mu_{i'}, \mu_{j'}\right)-\kappa*d_X\left(v^X_{i'}, v^X_{j}\right)\right|}{\partial \mu_{j}}
\end{aligned}
\end{equation*}

The Euclidean metric is used for both the input space and the hidden layer space, i.e., $d_Z\left(\mu_{i'}, \mu_{j'}\right)=\left\|\mu_{i'}-\mu_{j'}\right\|$. In addition, the symbols are somewhat abused for clear derivation, representing $\kappa*d_X\left(v^X_{i'}, v^X_{j'}\right)$ with $K$. Accordingly, the above equation can be further derived as follows:

\begin{equation*}
\begin{aligned}
& \frac{\partial L_{rank}}{\partial \mu_j}
 = \sum_{i'=1}^{C}\sum_{j'=1}^{C} \frac{\partial \left|d_Z\left(\mu_{i'}, \mu_{j'}\right)-\kappa*d_X\left(v^X_{i'}, v^X_{j'}\right)\right|}{\partial \mu_{j}}\\
& = \sum_{i'=1}^{C}\sum_{j'=1}^{C} \frac{\partial \big | \left\|\mu_{i'}-\mu_{j'}\right\|-K\big |}{\partial \mu_j}\\
& = \sum_{i'=1}^{C} \frac{\partial \big |\left\|\mu_{i'}-\mu_j\right\|-K\big |}{\partial \mu_j}+\sum_{j'=1}^{C} \frac{\partial \big |\left\|\mu_j-\mu_{j'}\right\|-K\big |}{\partial \mu_j}\\
& = \sum_{i'=1}^{C} \frac{\partial \left(\left\|\mu_{i'}-\mu_j\right\|-K\right)}{\partial \mu_j}\cdot\frac{\left\|\mu_{i'}-\mu_j\right\|-K}{\big |\left\|\mu_{i'}-\mu_j\right\|-K\big |}\\&\quad+\sum_{j'=1}^{C} \frac{\partial \left(\left\|\mu_j-\mu_{j'}\right\|-K\right)}{\partial \mu_j}\cdot\frac{\left\|\mu_j-\mu_{j'}\right\|-K}{\big |\left\|\mu_j-\mu_{j'}\right\|-K\big |}\\
& = \sum_{i'=1}^{C} \frac{\partial \left\|\mu_{i'}-\mu_j\right\|}{\partial \mu_j}\cdot\frac{\left\|\mu_{i'}-\mu_j\right\|-K}{\big |\left\|\mu_{i'}-\mu_j\right\|-K\big |}\\&\quad+\sum_{j'=1}^{C} \frac{\partial \left\|\mu_j-\mu_{j'}\right\|}{\partial \mu_j}\cdot\frac{\left\|\mu_j-\mu_{j'}\right\|-K}{\big |\left\|\mu_j-\mu_{j'}\right\|-K\big |}\\
& = \sum_{i'=1}^{C} \frac{\mu_j-\mu_{i'}}{\left\|\mu_j-\mu_{i'}\right\|}\cdot\frac{\left\|\mu_j-\mu_{i'}\right\|-K}{\big |\left\|\mu_j-\mu_{i'}\right\|-K\big |}\\&\quad+\sum_{j'=1}^{C} \frac{\mu_j-\mu_{j'}}{\left\|\mu_j-\mu_{j'}\right\|}\cdot\frac{\left\|\mu_j-\mu_{j'}\right\|-K}{\big |\left\|\mu_j-\mu_{j'}\right\|-K\big |}\\
& = 2\sum_{i'=1}^{C} \frac{\mu_j-\mu_{i'}}{\left\|\mu_j-\mu_{i'}\right\|}\cdot\frac{\left\|\mu_j-\mu_{i'}\right\|-K}{\big |\left\|\mu_j-\mu_{i'}\right\|-K\big |}\\
& = 2\sum_{i'=1}^{C} \frac{\mu_j-\mu_{i'}}{\left\|\mu_j-\mu_{i'}\right\|}\cdot\frac{\left\|\mu_j-\mu_{i'}\right\|-\kappa*d_X\left(v^X_{i'}, v^X_j\right)}{\big |\left\|\mu_j-\mu_{i'}\right\|-\kappa*d_X\left(v^X_{i'}, v^X_j\right)\big |}\\
& = 2\sum_{i'=1}^{C} \frac{\mu_j-\mu_{i'}}{d_Z\left(\mu_j, \mu_{i'}\right)}\cdot\frac{d_Z\left(\mu_j, \mu_{i'}\right)-\kappa*d_X\left(v^X_{i'}, v^X_j\right)}{\left|d_Z\left(\mu_j, \mu_{i'}\right)-\kappa*d_X\left(v^X_{i'}, v^X_j\right)\right|}\\
\end{aligned}
\end{equation*}

The gradient of $L_{align}$ with respect to each learnalbe cluster center $\mu_{j}$ can be computed as:

\begin{equation*}
\begin{aligned}
\frac{\partial L_{align}}{\partial \mu_j} 
& = \frac{\partial \sum_{j'=1}^C ||\mu_{j'}-v^Z_{j'}||}{\partial \mu_j}\\
& = \sum_{j'=1}^C \frac{\partial ||\mu_{j'}-v^Z_{j'}||}{\partial \mu_j}\\
& = \frac{\partial ||\mu_j-v^Z_{j}||}{\partial \mu_j} \\
& = \frac{\partial (\mu_j-v^Z_{j})}{\partial \mu_j} \cdot \frac{\mu_{j}-v_{j}^{Z}}{\left\|\mu_{j}-v_{j}^{Z}\right\|} \\
& = \frac{\mu_{j}-v_{j}^{Z}}{\left\|\mu_{j}-v_{j}^{Z}\right\|}
\end{aligned}
\end{equation*}

\begin{table*}[!htbp]
\begin{center}
\caption{Parameter Sensitivity with different parameters $k$ and $\kappa$ on the MNIST-test dataset.}
\label{tab:A2}
\begin{tabular}{cccccccc}
\hline
\textbf{Parameters} & \textbf{ACC/NMI}$\uparrow$ & \textbf{RRE}$\downarrow$ & \textbf{Trust}$\uparrow$ & \textbf{Cont}$\uparrow$ & \textbf{$d$-RMSE}$\downarrow$ & \textbf{LGD}$\downarrow$ & \textbf{CRA}$\uparrow$ \\ \hline
$k$=1, $\kappa$=3 & \textbf{0.975/0.936} & 0.0125 & 0.9944 & 0.9756 & 5.757 & \textbf{0.8868} & 1.00 \\
$k$=3, $\kappa$=3 & 0.973/0.931 & 0.0114 & 0.9970 & 0.9757 & 5.805 & 0.9207 & 1.00 \\
$k$=5, $\kappa$=3 & 0.972/0.930 & 0.0109 & 0.9981 & 0.9761 & 5.800 & 0.9339 & 1.00 \\
$k$=8, $\kappa$=3 & 0.972/0.929 & \textbf{0.0104} & 0.9989 & \textbf{0.9765} & 5.810 & 0.9476 & 1.00 \\
$k$=10, $\kappa$=3 & 0.972/0.929 & 0.0105 & \textbf{0.9990} & 0.9764 & \textbf{5.704} & 0.9487 & 1.00 \\ \hline
$k$=5, $\kappa$=1 & 0.967/0.912 & \textbf{0.0068} & \textbf{0.9993} & \textbf{0.9845} & \textbf{5.409} & \textbf{0.2524} & 1.00 \\
$k$=5, $\kappa$=3 & \textbf{0.972/0.930} & 0.0109 & 0.9981 & 0.9761 & 5.800 & 0.9339 & 1.00 \\
$k$=5, $\kappa$=5 & 0.972/0.929 & 0.0146 & 0.9964 & 0.9691 & 15.0653 & 1.5719 & 1.00 \\
$k$=5, $\kappa$=8 & 0.972/0.929 & 0.0190 & 0.9943 & 0.9615 & 29.4607 & 2.5410 & 1.00 \\
$k$=5, $\kappa$=10 & 0.972/0.929 & 0.0195 & 0.9951 & 0.9597 & 37.7661 & 3.1434 & 1.00 \\ \hline
\end{tabular}
\end{center}
\end{table*}

\section*{A.3 Definitions of Performance Metrics}
The following notations are used for the definitions:
\begin{itemize}
    \item[] $d_X(i,j)$: the pairwise distance between $x_i$ and $x_j$ in input space $X$; 
    \item[] $d_Z(i,j)$: the pairwise distance between $z_i$ and $z_j$ in latent space $Z$;
    \item[] $\mathcal{N}_{i}^{k,X}$: the set of indices to the $k$-nearest neighbor ($k$NN) of $x_i$ in input space $X$;
    \item[] $\mathcal{N}_{i}^{k,Z}$: the set of indices to the $k$-nearest neighbor ($k$NN) of $z_i$ in latent space $Z$;
    \item[] $r_X(i,j)$: the rank of the closeness (in Euclidean distance) of $x_j$ to $x_i$ in input space $X$; 
    \item[] $r_Z(i,j)$: the rank of the closeness (in Euclidean distance) of $z_j$ to $z_i$ in latent space $Z$.
\end{itemize}

The eight evaluation metrics are defined below:

\begin{enumerate}
    
    \item[(1)] \textbf{ACC} (Accuracy) measures the accuracy of clustering:
    
    \begin{align*}
     A C C=\max _{m} \frac{\sum_{i=1}^{N} 1\left\{l_{i}=m\left(s_{i}\right)\right\}}{N}
	\end{align*}
	
	where $l_i$ and $s_i$ are the true and predicted labels for data point $x_i$, respectively, and $m(\cdot)$ is all possible one-to-one mappings between clusters and label categories.
    
    \item[(2)] \textbf{NMI} (Normalized Mutual Information) NMI calculates the normalized measure of similarity between two labels of the same data
    
    \begin{align*}
     NMI=\frac{I(l ; s)}{\max \{H(l), H(s)\}}
	\end{align*}
	
	where $I(l,s)$ is the mutual information between the real label $l$ and predicted label $s$, and $H(\cdot)$ represents their entropy.
    
    \item[(3)] \textbf{RRE} (Relative Rank Change) measures the average of changes in neighbor ranking between two spaces $X$ and $Z$:
    
    \begin{align*}
     R R E=\frac{1}{\left(k_{2}-k_{1}+1\right)} \sum_{k=k_{1}}^{k_{2}}\left\{M R_{X \rightarrow Z}^{k}+M R_{Z \rightarrow X}^{k}\right\}
	\end{align*}
	
	where $k_1$ and $k_2$ are the lower and upper bounds of the $k$-NN.
	
    \begin{align*}
    M R_{X \rightarrow Z}^{k}=\frac{1}{H_{k}} \sum_{i=1}^{N} \sum_{j \in \mathcal{N}_{i}^{k,Z}}\left(\frac{\left|r_X(i, j)-r_Z(i, j)\right|}{r_Z(i, j)}\right)
    \end{align*}
    \begin{align*}
     M R_{Z \rightarrow X}^{k}=\frac{1}{H_{k}} \sum_{i=1}^{N} \sum_{j \in \mathcal{N}_{i}^{k,X}}\left(\frac{\left|r_X(i, j)-r_Z(i, j)\right|}{r_X(i, j)}\right)
    \end{align*}
    
	where $H_{k}$ is the normalizing term, defined as
	
    \begin{align*}
    H_{k}=N \sum_{l=1}^{k} \frac{|N-2 l|}{l}
    \end{align*}
 
    \item[(4)]  \textbf{Trust} (Trustworthiness)  measures to what extent the $k$ nearest neighbors of a point are preserved when going from the input space to the latent space:
    
    \begin{align*}
    {Trust} &=\frac{1}{k_{2}-k_{1}+1} \sum_{k=k_{1}}^{k_{2}}\left\{ 1-\frac{2}{N k(2 N-3 k-1)}\right. \\& \left. \quad\sum_{i=1}^{N} \sum_{j \in \mathcal{N}_{i}^{k,Z},j \notin \mathcal{N}_{i}^{k,X}}(r_X(i, j)-k)\right\}
    \end{align*}
    
	where $k_1$ and $k_2$ are the bounds of the number of nearest neighbors.
	 
    \item[(5)] \textbf{Cont} (Continuity) is defined analogously to $Trust$, but checks to what extent neighbors are preserved when going from the latent space to the input space:
    
    \begin{align*}
    {Cont} &=\frac{1}{k_{2}-k_{1}+1} \sum_{k=k_{1}}^{k_{2}}\left\{1-\frac{2}{N k(2 N-3 k-1)}\right. \\& \left. \quad\sum_{i=1}^{N} \sum_{j \notin \mathcal{N}_{i}^{k,Z},j \in \mathcal{N}_{i}^{k,X}}(r_Z(i, j)-k)\right\}
    \end{align*}
    
    where $k_1$ and $k_2$ are the bounds of the number of nearest neighbors.

    \item[(6)] \textbf{$d$-RMSE} (Root Mean Square Error) measures to what extent the two distributions of \textbf{distances} coincide:

    \begin{align*}
    d-RMSE=\sqrt{\frac{1}{N^2} \sum_{i=1}^{N}\sum_{j=1}^{N}\left(d_X(i,j)-d_Z(i,j)\right)^{2}}
	\end{align*}

    \item[(7)] \textbf{LGD} (Locally Geometric Distortion) measures how much corresponding distances between neighboring points differ in two metric spaces and is the primary metric for isometry, defined as:
    
    \begin{align*}
    LGD = \sum_{k=k_{1}}^{k_{2}} \sqrt{\sum_{i}^M \frac{ \sum_{j \in \mathcal{N}_{i}^{k,(l)}} (d_{l}(i, j)-d_{l'}(i, j))^2 }{  \left(k_{2}-k_{1}+1\right)^2  M(\#\mathcal{N}_{i}) }}
    \end{align*}
    
    where $k_1$ and $k_2$ are the lower and upper bounds of the $k$-NN.
    
    \item[(8)] \textbf{CRA} (Cluster Rank Accuracy) measures the changes in \textit{ranks} of cluster centers from the input space $X$ and to the latent space $Z$:
    
    \begin{align*}
    CRA=\frac{\sum_{i=1}^{C}\sum_{j=1}^{C} \textbf1(r_X(v^X_i, v^X_j)=r_Z(v^Z_i, v^Z_j))}{C^2}
    \end{align*}

    where $C$ is the number of clusters, $v^X_j$ is the cluster center of the $j$th cluster in the input space $X$, $v^Z_j$ is the cluster center of the $j$th cluster in the latent space $Z$, $r_X(v^X_i, v^X_j)$ denotes the rank of the closeness (in terms of Euclidean distance) of $v^X_i$ to $v^X_j$ in space $X$ in the input space $X$, and $r_Z(v^Z_i, v^Z_j)$ denotes the rank of the closeness (in terms of Euclidean distance) of $v^Z_i$ to $v^Z_j$ in space $Z$.
    
\end{enumerate}

\section*{A.4 Parameter Sensitivity}
We also evaluated the sensitivity of parameters $k$ and $\kappa$ on the MNIST-test dataset and the results are shown in Tab.~\ref{tab:A2}. It is found that the clustering performance is not sensitive to parameters $k$ and $\kappa$, and some combinations of $k$ and $\kappa$ even produce better clustering performance than the metrics reported in the main paper. However, the effect of $k$ and $\kappa$ on multi-manifold learning is more pronounced, and different combinations of $k$ and $\kappa$ may increase or decrease performance. In general, this paper focuses on the design of algorithms and implementations, and no hyperparameter search is performed to find the best performance metrics.

\section*{A.5 Evaluation of Multi-Manifold Learning}
GCML is compared with other methods on six manifold-related \emph{quantitative metrics} to demonstrate its performance advantage for multi-manifold learning, and the complete results on seven datasets are shown on the left side of Tab.~\ref{tab:A3}. In addition, the right side of Tab.~\ref{tab:A3} compares GCML with other methods on all datasets to see if these methods can really learn embeddings that are useful for downstream tasks. As shown in the table, GCML outperforms all the other methods with MLP, RFC, and LR as downstream tasks.

\section*{A.6 More Ablation Study Experiments}
The results of the ablation study on the MNIST-full dataset have been presented in Tab.~4 in Sec 4.4. Here, we provide complete ablation study results on all datasets in Tab.~\ref{tab:A4}. The conclusion is similar (note that the clustering performance of the model without clustering-oriented losses is very poorly, so the ``best" metric numbers are not meaningful and are marked in {\color[rgb]{0.7,0.7,0.7}gray} color): (1) CL is very important for obtaining good clustering. (2) SL is beneficial for both clustering and multi-manifold learning. (3) The training strategies (WC and AT) contribute to improving performance metrics such as RRE, Trust, Cont, and CRA.

\begin{table*}[!htbp]
	\begin{center}
	\caption{Performance for multi-manifold learning (left) and downstream tasks (right) on all seven datasets.}
	\label{tab:A3}
	\resizebox{\textwidth}{!}{
	\begin{tabular}{llcccccc|cccc}
		\hline
		\textbf{Datasets} &
		\multicolumn{1}{l}{\textbf{Algorithms}} &
		\multicolumn{1}{l}{\textbf{RRE}$\downarrow$} &
		\multicolumn{1}{c}{\textbf{Trust}$\uparrow$} &
		\multicolumn{1}{c}{\textbf{Cont}$\uparrow$} &
		\multicolumn{1}{c}{\textbf{$d$-RMSE}$\downarrow$} &
		\multicolumn{1}{c}{\textbf{LGD}$\downarrow$} &
		\multicolumn{1}{c}{\textbf{CRA}$\uparrow$} &
		\multicolumn{1}{c}{\textbf{MLP}$\uparrow$} &
		\multicolumn{1}{c}{\textbf{RFC}$\uparrow$} &
		\multicolumn{1}{c}{\textbf{SVM}$\uparrow$} &
		\multicolumn{1}{c}{\textbf{LR}$\uparrow$} \\ \hline
		\multirow{6}{*}{MNIST-full}   & DEC        & 0.09988          & 0.84499          & 0.94805          & 44.8535          & 4.37986          & 0.28   & 0.8647          & 0.8706          & 0.8707          & 0.8566      \\
		& IDEC       & 0.00984          & 0.99821          & 0.97936          & 24.5803          & 1.71484          & 0.33   & 0.9797          & 0.9737          & 0.9852          & 0.9650      \\
		& JULE       & 0.02657          & 0.93675          & 0.98321          & 28.3412          & 2.12955          & 0.27   & 0.9802          & 0.9825          & 0.9787          & 0.9743      \\
		& DSC        & 0.09785          & 0.87315          & 0.92508          & 6.98098          & 1.19886          & 0.23   & 0.9622          & 0.9501          & 0.9837          & 0.9752      \\
		& N2D        & 0.01002          & 0.99243          & 0.98466          & 5.7162           & 0.69946          & 0.21   & 0.9796          & 0.9803          & 0.9799          & 0.9792      \\
		& GCML (ours) & \textbf{0.00567} & \textbf{0.99978} & \textbf{0.98716} & \textbf{5.4986}  & \textbf{0.69168} & \textbf{1.00} & \textbf{0.9851} & \textbf{0.9874} & \textbf{0.9869} & \textbf{0.9841}\\ \hline
		\multirow{6}{*}{MNIST-test}   & DEC        & 0.12800          & 0.81841          & 0.91767          & 14.6113          & 2.29499          & 0.19    & 0.8525          & 0.8605          & 0.8725          & 0.8685     \\
		& IDEC       & 0.01505          & 0.99403          & 0.97082          & 7.4599           & 1.08350          & 0.38   & 0.9740          & 0.9725          & 0.9845          & 0.9655      \\
		& JULE       & 0.04122          & 0.92971          & 0.97208          & 9.4768           & 1.17176          & 0.42   & 0.9775          & 0.9845          & 0.9800          & 0.9825      \\
		& DSC        & 0.10728          & 0.85498          & 0.92254          & 7.1689           & 1.19239          & 0.26   & 0.9535          & 0.9740          & 0.9825          & 0.9795      \\
		& N2D        & 0.01565          & 0.98764          & 0.97572          & \textbf{5.0120}  & 0.97454          & 0.33   & 0.9715          & 0.9760          & 0.9725          & 0.9725      \\
		& GCML (ours) & \textbf{0.01090} & \textbf{0.99811} & \textbf{0.97612} & 5.8000           & \textbf{0.93394} & \textbf{1.00} & \textbf{0.9855} & \textbf{0.9875} & \textbf{0.9865} & \textbf{0.9855}\\ \hline
		\multirow{6}{*}{USPS}         & DEC        & 0.07911          & 0.88871          & 0.94628          & 16.4355          & 1.77848          & 0.31    & 0.8289          & 0.8668          & 0.8289          & 0.8294     \\
		& IDEC       & 0.01043          & 0.99726          & 0.97960          & 13.0573          & 1.11689          & 0.30   & 0.9482          & 0.9556          & 0.9656          & 0.9125       \\
		& JULE       & 0.02972          & 0.98763          & 0.98810          & 14.6324          & 1.43426          & 0.33   & 0.9576          & 0.9617          & \textbf{0.9703} & 0.9476      \\
		& DSC        & 0.06319          & 0.9151           & 0.93988          & 8.4412           & 1.02131          & 0.27   & 0.9351          & 0.9572          & 0.9612          & 0.9342      \\
		& N2D        & 0.01337          & 0.98769          & 0.98135          & 8.1961           & 0.54967          & 0.37   & 0.9569          & 0.9569          & 0.9569          & 0.9541      \\
		& GCML (ours) & \textbf{0.00577} & \textbf{0.99979} & \textbf{0.98701} & \textbf{6.4980}  & \textbf{0.53180} & \textbf{1.00} & \textbf{0.9656} & \textbf{0.9651} & 0.9604          & \textbf{0.9551}\\ \hline
		\multirow{6}{*}{Fashion-MNIST} & DEC        & 0.04787          & 0.93896          & 0.95450          & 39.3274          & 3.87731          & 0.37   & 0.6268          & 0.9853          & 0.6377          & 0.6245      \\
		& IDEC       & 0.01089          & 0.99683          & 0.97797          & 25.4024          & 1.91385          & 0.27      & 0.8367          & 0.9918          & \textbf{0.8607} & 0.7514   \\
		& JULE       & 0.03013          & 0.97732          & 0.97923          & 15.2213          & 1.43642          & 0.43      & 0.8541          & 0.9892          & 0.8566          & 0.7723   \\
		& DSC        & 0.05168          & 0.95013          & 0.96121          & 17.2201          & 1.42091          & 0.36      & 0.8084          & 0.9823          & 0.8618          & 0.7676   \\
		& N2D        & 0.00894          & 0.99062          & 0.98054          & 14.49079         & \textbf{1.28180} & 0.26      & 0.8412          & 0.9493          & 0.8230          & 0.7753   \\
		& GCML (ours) & \textbf{0.00836} & \textbf{0.99868} & \textbf{0.98203} & \textbf{13.3788} & 1.33893          & \textbf{1.00} & \textbf{0.8642} & \textbf{0.9942} & 0.8468          & \textbf{0.7768}\\ \hline
		\multirow{6}{*}{REUTERS-10K}  & DEC        & 0.26192          & 0.65518          & 0.80477          & 40.4671          & 4.00423          & 0.63    & 0.7985          & 0.7880          & 0.8105          & 0.7450     \\
		& IDEC       & 0.05981          & 0.95840          & 0.90550          & 43.9556          & 2.01365          & 0.75    & 0.9225          & 0.8930          & 0.9280          & 0.7705     \\
		& JULE       & -          & -          & -          & -          & -          & -        & -          & -          & -          & - \\
		& DSC        & -          & -          & -          & -          & -          & -        & -          & -          & -          & - \\
		& N2D        & 0.03827          & 0.97385          & 0.93412          & 36.1042          & \textbf{1.69013} & 0.31  & 0.9205          & 0.9080          & 0.9240          & 0.8335       \\
		& GCML (ours) & \textbf{0.03206} & \textbf{0.98380} & \textbf{0.93802} & \textbf{34.5478} & 2.72096          & \textbf{1.00} & \textbf{0.9360} & \textbf{0.9185} & \textbf{0.9390} & \textbf{0.8475}\\ \hline
		\multirow{6}{*}{HAR}  & DEC        & 0.09060          & 0.89097          & 0.91766          & 10.0222          & 1.58691          & 0.30   & 0.7696          & 0.7847          & 0.7628          & 0.7634      \\
		& IDEC       & 0.01031          & 0.99433          & 0.98132          & 9.9155          & 0.93736          & 0.39     & 0.8973          & 0.9031          & 0.9041          & 0.8822    \\
		& JULE       & -          & -          & -          & -          & -          & -    & -          & -          & -          & -     \\
		& DSC        & -          & -          & -          & -          & -          & -    & -          & -          & -          & -     \\
		& N2D        & 0.00841          & 0.99281          & 0.97695          & \textbf{8.2326}          & 0.64296 & 0.33    & 0.9138          & 0.9083          & 0.9174          & 0.8799     \\
		& GCML (ours) & \textbf{0.00665} & \textbf{0.99895} & \textbf{0.98634} & 15.2876 & \textbf{0.46189}  & \textbf{1.00} & \textbf{0.9235} & \textbf{0.9193} & \textbf{0.9293} & \textbf{0.8996}\\ \hline
		\multirow{6}{*}{pendigits}  & DEC        & 0.03932          & 0.95149          & 0.96300          & 21.6608          & 1.32970          & 0.24   & 0.7215          & 0.8432          & 0.7376          & 0.7429      \\
		& IDEC       & 0.00384          & 0.99879          & 0.99255          & 19.7243          & 0.86137          & 0.28     & 0.8870          & 0.9461          & \textbf{0.9595}          & 0.8636    \\
		& JULE       & -          & -          & -          & -          & -          & -    & -          & -          & -          & -     \\
		& DSC        & -          & -          & -          & -          & -          & -    & -          & -          & -          & -     \\
		& N2D        & 0.00262         & 0.99919          & 0.99473          & 20.7052          & 0.76941 & 0.42    & 0.9131          & 0.9669          & 0.9528          & 0.8516     \\
		& GCML (ours) & \textbf{0.00091} & \textbf{0.99994} & \textbf{0.99808} & \textbf{2.5184} & \textbf{0.08223}  & \textbf{1.00} & \textbf{0.9538} & \textbf{0.9705} & 0.9532 & \textbf{0.8953}\\ \hline
	\end{tabular}}
	\end{center}
\end{table*}
\begin{table*}[!htbp]
\begin{center}
\caption{Ablation study of loss items and training strategies used in the proposed GCML framework.}
\label{tab:A4}
\begin{tabular}{cllllllll}
\hline
\textbf{Datasets} &
  \multicolumn{1}{c}{\textbf{Methods}} &
  \multicolumn{1}{c}{\textbf{ACC/NMI}$\uparrow$} &
  \multicolumn{1}{c}{\textbf{RRE}$\downarrow$} &
  \multicolumn{1}{c}{\textbf{Trust}$\uparrow$} &
  \multicolumn{1}{c}{\textbf{Cont}$\uparrow$} &
  \textbf{$d$-RMSE}$\downarrow$ &
  \textbf{LGD}$\downarrow$ &
  \textbf{CRA}$\uparrow$ \\ \hline
 &
  w/o SL &
  0.976/0.939 &
  0.0093 &
  0.9967 &
  0.9816 &
  24.589 &
  1.6747 &
  0.32 \\
 &
  {\color[rgb]{0.75,0.75,0.75} w/o CL} &
  {\color[rgb]{0.75,0.75,0.75} 0.814/0.736} &
  {\color[rgb]{0.75,0.75,0.75} 0.0004} &
  {\color[rgb]{0.75,0.75,0.75} 0.9998} &
  {\color[rgb]{0.75,0.75,0.75} 0.9990} &
  {\color[rgb]{0.75,0.75,0.75} 7.458} &
  {\color[rgb]{0.75,0.75,0.75} 0.0487} &
  {\color[rgb]{0.75,0.75,0.75} 1.00} \\
 &
  w/o WC &
  0.977/0.943 &
  0.0065 &
  0.9987 &
  0.9860 &
  5.576 &
  0.6968 &
  0.98 \\
 &
  w/o AT &
  0.978/0.944 &
  0.0069 &
  0.9986 &
  0.9851 &
  5.617 &
  0.7037 &
  0.96 \\
\multirow{-5}{*}{MNIST-full} &
  full model &
  \textbf{0.980/0.946} &
  \textbf{0.0056} &
  \textbf{0.9997} &
  \textbf{0.9871} &
  \textbf{5.498} &
  \textbf{0.6916} &
  \textbf{1.00} \\ \hline
 &
  w/o SL &
  \textbf{0.973/0.932} &
  0.0146 &
  0.9928 &
  0.9727 &
  7.701 &
  1.0578 &
  0.31 \\
 &
  {\color[rgb]{0.75,0.75,0.75} w/o CL} &
  {\color[rgb]{0.75,0.75,0.75} 0.773/0.747} &
  {\color[rgb]{0.75,0.75,0.75} 0.0020} &
  {\color[rgb]{0.75,0.75,0.75} 0.9994} &
  {\color[rgb]{0.75,0.75,0.75} 0.9954} &
  {\color[rgb]{0.75,0.75,0.75} 7.229} &
  {\color[rgb]{0.75,0.75,0.75} 0.0809} &
  {\color[rgb]{0.75,0.75,0.75} 1.00} \\
 &
  w/o WC &
  0.956/0.904 &
  0.0132 &
  0.9955 &
  0.9735 &
  \textbf{5.470} &
  0.9364 &
  \textbf{1.00} \\
 &
  w/o AT &
  0.970/0.929 &
  0.0118 &
  0.9974 &
  0.9747 &
  5.567 &
  0.9404 &
  \textbf{1.00} \\
\multirow{-5}{*}{MNIST-test} &
  full model &
  0.972/0.930 &
  \textbf{0.0109} &
  \textbf{0.9981} &
  \textbf{0.9761} &
  5.800 &
  \textbf{0.9339} &
  \textbf{1.00} \\ \hline
 &
  w/o SL &
  0.957/0.902 &
  0.0095 &
  0.9967 &
  0.9812 &
  14.609 &
  0.9847 &
  0.29 \\
 &
  {\color[rgb]{0.75,0.75,0.75} w/o CL} &
  {\color[rgb]{0.75,0.75,0.75} 0.664/0.658} &
  {\color[rgb]{0.75,0.75,0.75} 0.0020} &
  {\color[rgb]{0.75,0.75,0.75} 0.9996} &
  {\color[rgb]{0.75,0.75,0.75} 0.9952} &
  {\color[rgb]{0.75,0.75,0.75} 2.934} &
  {\color[rgb]{0.75,0.75,0.75} 0.0687} &
  {\color[rgb]{0.75,0.75,0.75} 1.0} \\
 &
  w/o WC &
  0.956/0.896 &
  0.0060 &
  0.9991 &
  0.9868 &
  6.572 &
  0.5335 &
  \textbf{1.00} \\
 &
  w/o AT &
  0.947/0.885 &
  0.0080 &
  0.9979 &
  0.9833 &
  \textbf{5.960} &
  \textbf{0.4967} &
  \textbf{1.00} \\
\multirow{-5}{*}{USPS} &
  full model &
  \textbf{0.958/0.902} &
  \textbf{0.0057} &
  \textbf{0.9997} &
  \textbf{0.9870} &
  6.498 &
  0.5318 &
  \textbf{1.00} \\ \hline
 &
  w/o SL &
  0.706/0.682 &
  0.0108 &
  0.9964 &
  0.9781 &
  25.954 &
  1.8936 &
  0.30 \\
 &
  {\color[rgb]{0.75,0.75,0.75} w/o CL} &
  {\color[rgb]{0.75,0.75,0.75} 0.576/0.569} &
  {\color[rgb]{0.75,0.75,0.75} 0.0004} &
  {\color[rgb]{0.75,0.75,0.75} 0.9994} &
  {\color[rgb]{0.75,0.75,0.75} 0.9995} &
  {\color[rgb]{0.75,0.75,0.75} 7.654} &
  {\color[rgb]{0.75,0.75,0.75} 0.0523} &
  {\color[rgb]{0.75,0.75,0.75} 1.00} \\
 &
  w/o WC &
  0.702/0.695 &
  0.0084 &
  0.9972 &
  0.9814 &
  \textbf{13.238} &
  1.3474 &
  \textbf{1.00} \\
 &
  w/o AT &
  0.708/0.694 &
  0.0097 &
  0.9975 &
  0.9798 &
  13.354 &
  1.3611 &
  \textbf{1.00} \\
\multirow{-5}{*}{Fashion-MNIST} &
  full model &
  \textbf{0.710/0.685} &
  \textbf{0.0083} &
  \textbf{0.9986} &
  \textbf{0.9820} &
  13.378 &
  \textbf{1.3389} &
  \textbf{1.00} \\ \hline
 &
  w/o SL &
  0.819/0.564 &
  0.0529 &
  0.9610 &
  0.9185 &
  44.481 &
  1.9090 &
  0.38 \\
 &
  {\color[rgb]{0.75,0.75,0.75} w/o CL} &
  {\color[rgb]{0.75,0.75,0.75} 0.542/0.279} &
  {\color[rgb]{0.75,0.75,0.75} 0.0277} &
  {\color[rgb]{0.75,0.75,0.75} 0.9868} &
  {\color[rgb]{0.75,0.75,0.75} 0.9456} &
  {\color[rgb]{0.75,0.75,0.75} 37.018} &
  {\color[rgb]{0.75,0.75,0.75} 2.2294} &
  {\color[rgb]{0.75,0.75,0.75} 1.00} \\
 &
  w/o WC &
  0.830/0.583 &
  0.0420 &
  0.9667 &
  0.9361 &
  35.302 &
  2.8286 &
  \textbf{1.00} \\
 &
  w/o AT &
  0.825/0.563 &
  0.0440 &
  0.9650 &
  0.9330 &
  39.275 &
  2.9146 &
  \textbf{1.00} \\
\multirow{-5}{*}{REUTERS-10K} &
  full model &
  \textbf{0.836/0.590} &
  \textbf{0.0320} &
  \textbf{0.9838} &
  \textbf{0.9380} &
  \textbf{34.547} &
  \textbf{2.7209} &
  \textbf{1.00} \\ \hline
  &
  w/o SL &
  0.835/0.746 &
  0.0116 &
  0.9944 &
  0.9792 &
  \textbf{8.168} &
  0.8882 &
  0.33 \\
 &
  {\color[rgb]{0.75,0.75,0.75} w/o CL} &
  {\color[rgb]{0.75,0.75,0.75} 0.744/0.615} &
  {\color[rgb]{0.75,0.75,0.75} 0.0024} &
  {\color[rgb]{0.75,0.75,0.75} 0.9986} &
  {\color[rgb]{0.75,0.75,0.75} 0.9948} &
  {\color[rgb]{0.75,0.75,0.75} 15.060} &
  {\color[rgb]{0.75,0.75,0.75} 0.2193} &
  {\color[rgb]{0.75,0.75,0.75} 1.00} \\
 &
  w/o WC &
  0.786/0.701 &
  0.0130 &
  0.9950 &
  0.9756 &
  15.398 &
  0.6171 &
  \textbf{1.00} \\
 &
  w/o AT &
  0.834/0.745 &
  0.0089 &
  0.9965 &
  0.9835 &
  15.726 &
  0.4734 &
  \textbf{1.00} \\
\multirow{-5}{*}{HAR} &
  full model &
  \textbf{0.844/0.862} &
  \textbf{0.0066} &
  \textbf{0.9989} &
  \textbf{0.9863} &
  15.287 &
  \textbf{0.4618} &
  \textbf{1.00} \\ \hline
  &
  w/o SL &
  0.843/0.803 &
  0.0027 &
  0.9989 &
  0.9948 &
  20.692 &
  0.7587 &
  0.31 \\
 &
  {\color[rgb]{0.75,0.75,0.75} w/o CL} &
  {\color[rgb]{0.75,0.75,0.75} 0.773/0.705} &
  {\color[rgb]{0.75,0.75,0.75} 0.0004} &
  {\color[rgb]{0.75,0.75,0.75} 0.9998} &
  {\color[rgb]{0.75,0.75,0.75} 0.9992} &
  {\color[rgb]{0.75,0.75,0.75} 2.177} &
  {\color[rgb]{0.75,0.75,0.75} 0.0385} &
  {\color[rgb]{0.75,0.75,0.75} 1.00} \\
 &
  w/o WC &
  0.840/0.790 &
  0.0014 &
  0.9995 &
  0.9972 &
  2.690 &
  0.1124 &
  \textbf{1.00} \\
 &
  w/o AT &
  0.842/0.804 &
  0.0018 &
  0.9991 &
  0.9965 &
  2.756 &
  0.1148 &
  \textbf{1.00} \\
\multirow{-5}{*}{pendigits} &
  full model &
  \textbf{0.855/0.814} &
  \textbf{0.0009} &
  \textbf{0.9999} &
  \textbf{0.9981} &
  \textbf{2.518} &
  \textbf{0.0822} &
  \textbf{1.00} \\ \hline
\end{tabular}
\end{center}
\end{table*}

\end{document}